\documentclass[lettersize,journal]{IEEEtran}
\usepackage{amsmath,amsfonts}
\usepackage{algorithmic}
\usepackage{algorithm}
\usepackage{array}
\usepackage[caption=false,font=normalsize,labelfont=sf,textfont=sf]{subfig}
\usepackage{textcomp}
\usepackage{stfloats}
\usepackage{url}
\usepackage{verbatim}
\usepackage{graphicx}
\usepackage{cite}
\usepackage[hidelinks]{hyperref}
\hypersetup{pdftitle={Rotation Equivariant Mamba for Vision Tasks}, pdfauthor={Zhongchen Zhao, Qi Xie, Keyu Huang, Lei Zhang, Deyu Meng, and Zongben Xu}}

\usepackage{nicefrac}       
\usepackage{epsfig}
\usepackage{graphicx}
\usepackage{amssymb}
\usepackage{mathrsfs}
\usepackage{url}
\usepackage{amsthm}
\usepackage{bm}
\usepackage{footmisc}
\usepackage{color}
\usepackage{xfrac}
\usepackage[T1]{fontenc}
\usepackage{pifont}
\usepackage{units}

\usepackage[utf8]{inputenc}
\usepackage{adjustbox}
\usepackage{multirow}
\usepackage{multicol}
\usepackage{makecell}
\usepackage{colortbl}
\usepackage{xcolor}
\usepackage{booktabs}

\def\l{\left}
\def\r{\right}

\def\({\l(}
\def\){\r)}
\def\[{\l[}
\def\]{\r]}

\usepackage{booktabs}
\usepackage{wrapfig}
\usepackage{xcolor}
\definecolor{deepred}{HTML}{C00000}

\newtheorem{Theorem}{Theorem}

\newtheorem{Lemma}{Lemma}

\newcommand{\piEQ}[2]{\pi^{\scriptscriptstyle #1}_{\scriptscriptstyle #2}}

\begin{document}
	
	\title{Rotation Equivariant Mamba for Vision Tasks}
	
\author{
	Zhongchen Zhao, Qi Xie, Keyu Huang, Lei Zhang, Deyu Meng, and Zongben Xu
	\thanks{This work was supported in part by the National Key R\&D Program of China under Grant 2024YFA1012000; in part by the China NSFC projects under Grant 12622132; in part by the Fundamental and Interdisciplinary Disciplines Breakthrough Plan of the Ministry of Education of China (No. JYB2025XDXM101); in part by the China NSFC project 62476214; and in part by the Tianyuan Fund for Mathematics of the National Natural Science Foundation of China (Grant No. 12426105) \textit{(Corresponding author: Qi Xie.)}}
	\thanks{Zhongchen Zhao, Qi Xie, Keyu Huang, and Zongben Xu are with the School of Mathematics and Statistics, Xi'an Jiaotong University, Shaanxi, P.R. China (e-mail: zhongchenzhao@stu.xjtu.edu.cn, xie.qi@mail.xjtu.edu.cn, keyuhuang@stu.xjtu.edu.cn, zbxu@mail.xjtu.edu.cn).}
	\thanks{Deyu Meng is with the School of Mathematics and Statistics and Ministry of Education Key Lab of Intelligent Networks and Network Security, Xi'an Jiaotong University, Xi'an, Shaanxi, China, and Macao Institute of Systems Engineering, Macau University of Science and Technology, Taipa, Macao (e-mail: dymeng@mail.xjtu.edu.cn).}
	\thanks{Lei Zhang is with the Department of Computing, the Hong Kong Polytechnic University, Hong Kong, P.R. China (e-mail: cslzhang@comp.polyu.edu.hk).}
}
	
	\markboth{Journal of \LaTeX\ Class Files,~Vol.~14, No.~8, April~2026}%
	{Zhao \MakeLowercase{\textit{et al.}}: Rotation Equivariant Mamba for Vision Tasks}
	
	\maketitle
	
	\begin{abstract}
		
		Rotation equivariance {is an important structural prior for a broad range of vision tasks, particularly in domains such as remote sensing and medical imaging where objects have no canonical orientation,}
		yet it remains notably absent from current Mamba-based vision architectures. Despite the success of Mamba in natural language processing and its growing adoption in computer vision, existing visual Mamba models fail to account for rotational symmetry in their design. This omission renders them sensitive to image rotations, thereby constraining their robustness and cross-task generalization. To address this limitation, we incorporate rotation symmetry, 
		{an important geometric prior in many visual domains},
		into Mamba-based architectures. Specifically, we introduce EQ-VMamba, the first rotation equivariant visual Mamba architecture for vision tasks. The core components of EQ-VMamba include a carefully designed rotation equivariant cross-scan strategy and group Mamba blocks. Moreover, we provide a rigorous theoretical analysis of the intrinsic equivariance error, demonstrating that the proposed architecture enforces end-to-end rotation equivariance {under the $\mathrm{p}4$ rotation group}. Extensive experiments across multiple benchmarks---including high-level image classification, mid-level semantic segmentation, and low-level image super-resolution---demonstrate that EQ-VMamba consistently improves rotation robustness and {achieves superior or competitive performance compared to non-equivariant baselines on most evaluated benchmarks}, while requiring 40--70\% fewer parameters.
		These results indicate that embedding rotation equivariance not only effectively bolsters the robustness of visual Mamba models against rotation transformations, but 
		{also improves accuracy on data exhibiting rotational symmetry, with substantially improved parameter efficiency.}
		Code is available at \url{https://github.com/zhongchenzhao/EQ-VMamba}.
	\end{abstract}
	

	\section{Introduction}
	\IEEEPARstart{I}{n} recent years, Mamba~\cite{gu2024mamba} has emerged as one of the most promising next-generation foundational architectures in deep learning, succeeding convolutional neural networks (CNNs)~\cite{simonyan2014very,resnet,huang2017densely,tan2019efficientnet,liu2022convnet} and Transformers~\cite{vaswani2017attention,vit,liu2021swin}. Unlike Transformers, which suffer from quadratic computational complexity with respect to input length, Mamba builds upon the core paradigm of State Space Models (SSMs)~\cite{gu2021combining,gu2021efficiently,smith2022simplified} and introduces an interpretable selective scan mechanism to model long-range dependencies, establishing itself as the first linear-complexity architecture to achieve performance comparable to Transformers in natural language processing (NLP).
	
	Following Mamba's success in NLP, numerous studies~\cite{zhu2024ViM,2024Vmamba,SpatialMamba,zhao2025polyline,2024Mambair} have explored its adaptation to the visual domain. For instance, Vim~\cite{zhu2024ViM} directly replaces the self-attention mechanism in Vision Transformers (ViTs) with a bidirectional Mamba block, providing the first validation of Mamba's feasibility for vision applications. To better accommodate the structural discrepancy between 2D image tokens and 1D textual sequences, VMamba~\cite{2024Vmamba} introduces a Visual State-Space (VSS) block with a cross-scan strategy, enabling each token to symmetrically integrate global information from four directions in the 2D plane. As a result, VMamba surpasses many Transformer-based networks, including the representative Swin Transformer~\cite{liu2021swin}, on high-level vision tasks, highlighting the potential of Mamba-based architectures for visual recognition. Subsequently, MambaIR~\cite{2024Mambair} extends this cross-scan strategy to low-level vision tasks, proposing the first Mamba-based model for image restoration and achieving comparable or superior results to state-of-the-art Transformer-based models.
	
	\begin{figure}[t]
		\centering
		\includegraphics[width=0.49\textwidth]{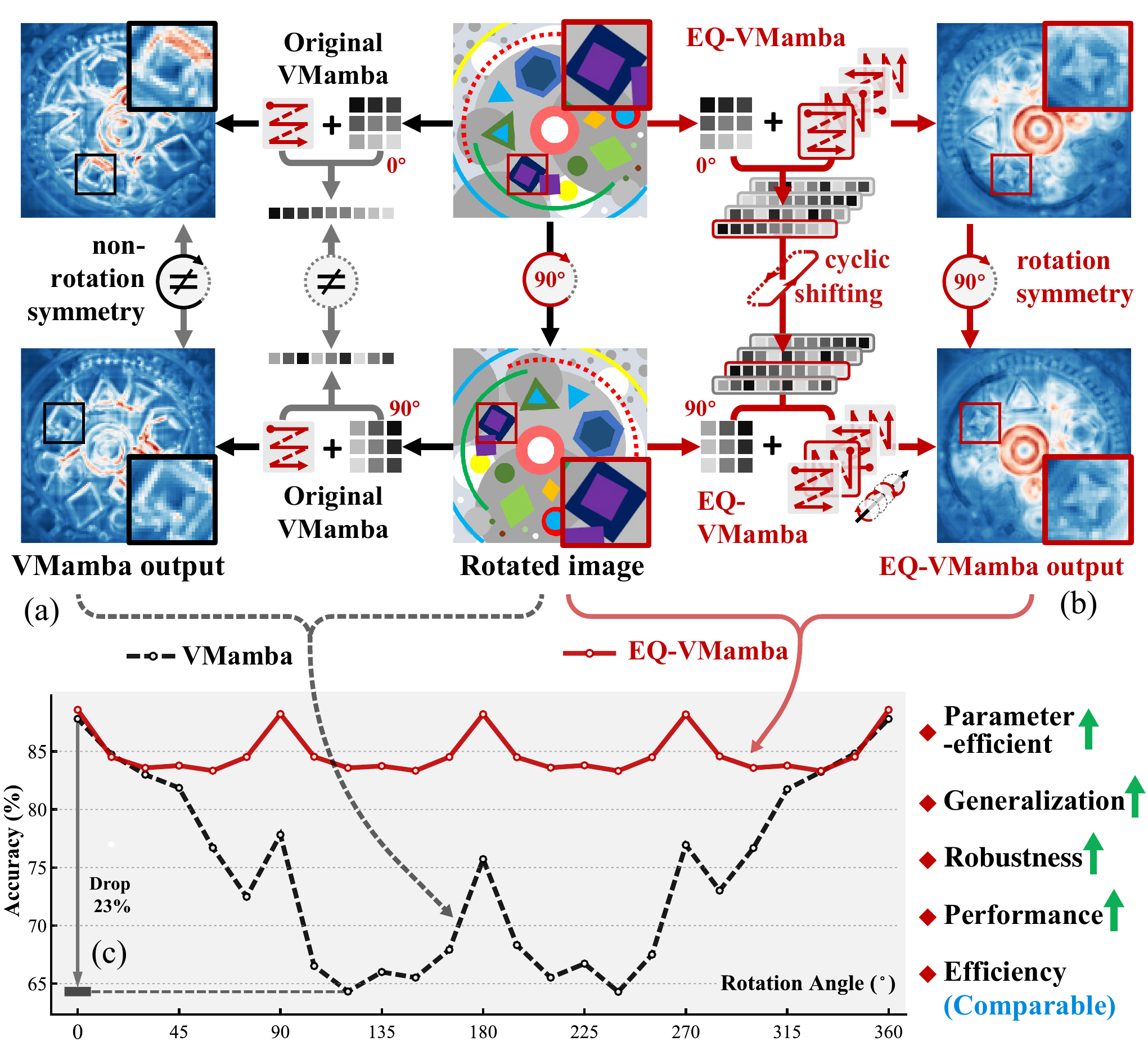}
		\vspace{-6mm}
		\caption{(a)-(b) Visualization of output feature maps for an input image and its rotated version using VMamba and our proposed EQ-VMamba, respectively. (c) Performance comparison between VMamba and EQ-VMamba on the rotated ImageNet-100 dataset.
		}
		\label{fig:visualization_EQ_Mamba}
		\vspace{-5mm}
	\end{figure}

	Despite these advances, directly adapting Mamba from NLP to vision remains far from fully exploiting its potential, due to the fundamental cross-modal discrepancies. Existing studies, represented by VMamba~\cite{2024Vmamba},  have primarily focused on bridging the structural gaps between images and language, such as the mismatch between 1D sequences and 2D spatial layouts. 
	However, they largely overlook another critical issue: the discrepancies in intrinsic content properties. Specifically, unlike textual data, natural images exhibit pronounced geometric symmetries: identical or similar patterns recur across various positions, orientations, and scales within an image~\cite{zeiler2014visualizing}. Neglecting these symmetries inherently limits the design efficiency of Mamba-based vision models, hindering the pursuit of more parameter-efficient architectures.

	In recent years, incorporating geometric symmetry priors into neural networks has been established as an effective approach to enhance generalization and robustness~\cite{cohen2016group,weiler2019general,shen2020pdoeconvs,2022Fconv,he2021efficient,xu20232,kundu2024steerable,cohen2016steerable,worrall2017harmonic,zhou2017oriented,marcos2017rotation,weiler2018learning,romero2020group,2025Bconv,bai2025regularization,2025spectralvmamba}. Formally, prior works~\cite{cohen2016group,weiler2019general} have formalized such priors as the requirement of equivariance: a model's prediction for a transformed image (e.g., translated, rotated, or reflected) should be consistent with the transformation applied to the prediction for the original image. Fundamental architectures such as CNNs and ViTs inherently possess certain equivariant properties. For instance, CNNs explicitly embed translation equivariance through weight sharing in sliding convolution kernels; rotation equivariant CNNs~\cite{cohen2016group,weiler2019general} further extend this by sharing kernels across an additional group dimension, thereby reducing the number of parameters while improving generalization. Similarly, recent studies~\cite{romero2020group,he2021efficient,xu20232,kundu2024steerable,hutchinson2021lietransformer,2025Bconv} reveal that the self-attention mechanism in ViTs exhibits inherent translation, rotation, and reflection equivariances, enabling these models to avoid interference from redundant spatial transformations and efficiently extract semantically meaningful features. 
	
	{In contrast to CNNs and ViTs, for which rotation equivariant counterparts have been well established, VMamba does not guarantee rotation equivariance by construction.}
	This limitation primarily stems from its core Visual State-Space (VSS) block, which exhibits weaker rotation equivariance than the convolution operators in CNNs and the self-attention mechanism in ViTs.
	Specifically, although the VSS block employs a four-way scanning mechanism to traverse 2D image tokens, rotational transformations of the input image still undesirably alter the scanning order, rendering the model highly sensitive to such transformations at the global level. 
	{Moreover, unlike learnable operators in CNNs and ViTs, whose equivariance can be enforced by constraining the learnable weights, the predefined and fixed scanning mechanism in VMamba offers no such handle, so that rotation equivariance is not guaranteed by construction.}
	Furthermore, the rotational symmetry of local structures is also disrupted during the tokenization process and the feature modeling within VSS blocks.  
	
	This lack of equivariance, spanning both global and local levels, results in poor robustness of VMamba against rotational perturbations across high-level, mid-level, and low-level vision tasks.
	As illustrated in Fig.~\ref{fig:visualization_EQ_Mamba}, when the same image is presented at different rotation angles, VMamba extracts different features and produces entirely different outputs (Fig.~\ref{fig:visualization_EQ_Mamba}(a)), leading to significant performance degradation on both image classification (Fig.~\ref{fig:visualization_EQ_Mamba}(c)) and semantic segmentation (Table~\ref{tab:rotated_semantic_segmentation_results}).
	These observations underscore the importance of developing a rotation equivariant VMamba to enhance the model's generalization, robustness, and parameter efficiency.

	However, designing a strictly rotation equivariant VMamba presents several non-trivial challenges.
	{Following the design principle of group-equivariant networks, we guarantee end-to-end rotation equivariance by making each module within the architecture individually equivariant, since equivariance is preserved under composition.}
	This imposes three key requirements:
	1) a redesigned rotation equivariant scanning mechanism for 2D images;
	2) a reformulation of the Mamba block to preserve equivariance throughout the state-space transformation;
	3) the adaptation of other non-equivariant modules (e.g., the patch embedding layer and task-specific decoders) to rotation equivariant counterparts.

	 While recent works recognize the critical importance of rotation robustness for VMamba, they often fail to confront these structural challenges directly. 
	 For instance, Spectral VMamba~\cite{2025spectralvmamba} provides a compromised solution by adopting orientation-agnostic local feature extraction to {achieve a rotation \textit{invariant} model\footnote{{Invariance requires that the output remains unchanged under input rotations. 
	 In contrast, equivariance is a more general property, as invariance can always be recovered from an equivariant representation, but not conversely. 
	 However, achieving invariance is considerably simpler than achieving equivariance, as the latter requires directional information to be rigorously preserved throughout the entire network. }}.
	 Furthermore, discarding the orientation information of local features inevitably restricts the model's applicability and leads to significant performance degradation, especially in dense prediction tasks (e.g., semantic segmentation and low-level vision) where the output must spatially align with the input. 
	 Consequently, establishing a fully \textit{equivariant} visual Mamba architecture remains a critical yet unresolved necessity.}

	In this paper, we propose EQ-VMamba, a strictly 90-degree rotation equivariant visual Mamba architecture that effectively alleviates the poor robustness of VMamba to image rotations.
	To the best of our knowledge, this is the first work to extend rotation equivariant design principles from CNN- and Transformer-based architectures to the emerging Mamba-based architecture, thereby enriching the existing framework of equivariant neural networks. 
	The main contributions of this paper can be summarized as follows:

	\begin{itemize}
		\item \textbf{Model Architecture:} We present the first rigorous formulation for designing a 90-degree rotation equivariant Mamba architecture.
		{Specifically, we propose a rotation equivariant cross-scan (EQ-cross-scan) strategy, whose four rotationally symmetric paths keep the scanning order in exact correspondence with the spatial rotation.
		Furthermore, we reformulate the Mamba block into a rotation equivariant group Mamba block, in which the input-dependent parameters are generated and transformed jointly with their corresponding tokens under the p4 rotation.}
		Building on these two designs, we construct an equivariant Visual State-Space (EQ-VSS) block and adapt two representative models, VMamba for high/mid-level and MambaIR for low-level vision tasks, into their equivariant counterparts: EQ-VMamba and EQ-MambaIR.
		
		\item \textbf{Parameter Efficiency:} Through careful architectural design, EQ-VMamba maintains a computational complexity comparable to its non-equivariant counterparts.
		Moreover, by sharing parameters across the group dimension, our approach reduces the total number of learnable parameters by approximately 50\% (as shown in Table~\ref{tab1:imagenet}), significantly enhancing parameter utilization efficiency.
		
		\item \textbf{Theoretical Analysis:} {We provide a rigorous equivariance guarantee for the proposed EQ-VMamba architecture.}
		Through the equivariance error analysis, we prove that the EQ-cross-scan/merge, the group Mamba blocks, and the overall EQ-VMamba architecture all achieve zero equivariance error under 90-degree rotations.
		
		\item \textbf{Empirical Validation:} Extensive experiments across high-level image classification, mid-level semantic segmentation, and low-level super-resolution tasks demonstrate that EQ-VMamba and EQ-MambaIR 
		{achieve accuracy comparable or superior to their non-equivariant counterparts, with the largest gains on data without a canonical orientation such as remote sensing and medical imagery.}
		Evaluations on rotated benchmarks confirm that EQ-VMamba effectively alleviates the rotational robustness limitations of visual Mamba.
		Collectively, these results demonstrate that incorporating rotation equivariance into the Mamba framework not only significantly enhances robustness against rotation transformations 
		{but also improves accuracy in most evaluated settings}.
	\end{itemize}
	
	The remainder of this paper is organized as follows.
	Sec.~\ref{sec:Related_Work} reviews related work.
	Sec.~\ref{sec:Method} details the proposed rotation equivariant visual Mamba, together with theoretical analyses of its equivariance error.
	Sec.~\ref{sec:Experiments} presents extensive experimental results across multiple vision tasks.
	Finally, Sec.~\ref{sec:Conclusion} concludes the paper and discusses future research directions.
	
	\section{Related Work}\label{sec:Related_Work}
	\subsection{Visual Mamba}
	As the dominant foundational architecture in deep learning, the Transformer~\cite{vaswani2017attention} is primarily constrained by the quadratic complexity of its self-attention mechanism, which scales quadratically with input sequence length.
	This computational bottleneck severely restricts its efficiency when modeling long-range dependencies.
	In recent years, numerous sub-quadratic architectures have been proposed to mitigate this limitation, including linear attention~\cite{katharopoulos2020transformers}, sparse Transformers~\cite{child2019generating}, and various recurrent neural networks~\cite{hochreiter1997long, peng2023rwkv, sun2023retentive}.
	However, these alternatives generally struggle to strike an optimal balance between computational efficiency and modeling capacity.
	
	Recently, Mamba~\cite{gu2024mamba}, a representative state space model (SSM), has for the first time achieved Transformer-level performance on NLP tasks while maintaining linear-time complexity, marking a significant milestone in efficient sequence modeling.
	Specifically, Mamba inherits the recurrent propagation mechanism of SSMs~\cite{gu2021efficiently,gu2021combining,smith2022simplified} for linear-time sequence modeling, and further introduces a powerful input-dependent selective scan mechanism, allowing the model to selectively propagate or forget historical  information along the scanning sequence.
	Owing to its efficiency and efficacy, Mamba has rapidly emerged as a promising alternative to Transformers, attracting considerable attention from the research community.
	
	Inspired by Mamba's success in NLP, numerous studies~\cite{zhu2024ViM,2024Vmamba,huang2024localmamba,shi2024multi,SpatialMamba,zhao2025polyline,han2024demystify} have explored its applicability to vision tasks.
	Vim~\cite{zhu2024ViM} pioneered this by replacing the original unidirectional scanning strategy with a bidirectional counterpart tailored for images.
	Subsequently, VMamba~\cite{2024Vmamba} observed that the 1D scanning strategy originally designed for text sequences in Mamba is ill-suited to the 2D structure of images and proposed a cross-scan strategy (i.e., a four-way scanning mechanism) to integrate contextual global information from four directions.
	Coupled with a series of architectural refinements, VMamba achieves superior performance on high-level vision tasks, surpassing Swin Transformer~\cite{liu2021swin} in both accuracy and inference throughput.

	Beyond high-level vision tasks, Mamba has been extended to a broader range of visual domains, including image restoration~\cite{2024Mambair,guo2025mambairv2,zou2024freqmamba}, image generation~\cite{teng2024dim}, video analysis~\cite{li2024videomamba}, remote sensing images~\cite{zhu2024samba,zhao2024rs}, point cloud analysis~\cite{liang2024pointmamba,zhang2024voxel}, and medical image analysis~\cite{ruan2024vm}.
	Among these, MambaIR~\cite{2024Mambair} introduced the first Mamba-based framework for image restoration, leveraging the cross-scan strategy of VMamba to enhance local features with a global receptive field while incorporating channel attention mechanisms to mitigate channel redundancy.
	Extensive experiments on various image restoration benchmarks demonstrate that MambaIR achieves competitive or superior performance compared to many well-established CNN- and Transformer-based methods~\cite{liang2021swinir,zhou2023srformer}.

	Although the significant potential of Mamba-based architectures in the visual domain has been demonstrated, 
	visual Mamba still suffers from an evident limitation: poor robustness to rotation transformations of input images.
	To address this deficiency, we develop EQ-VMamba by explicitly embedding rotation equivariance into the Mamba architecture.

	\subsection{Rotation Equivariant Neural Networks}
	Incorporating rotation symmetry priors into neural networks has long been an important research direction in deep learning.
	Over the past few years, numerous studies~\cite{cohen2016group,weiler2019general,shen2020pdoeconvs,2022Fconv,he2021efficient,xu20232,kundu2024steerable,cohen2016steerable,worrall2017harmonic,zhou2017oriented,marcos2017rotation,weiler2018learning,romero2020group,2025Bconv,bai2025regularization,2025spectralvmamba} have attempted to explicitly or implicitly incorporate the rotation symmetry prior into various network architectures.
	Early efforts~\cite{laptev2016ti,sohn2012learning,marcos2017rotation,worrall2017harmonic} primarily relied on data augmentation~\cite{krizhevsky2012imagenet} and regularization to implicitly enforce rotation equivariance, but these methods lack formal theoretical guarantees and often exhibit unstable performance in preserving strict equivariance. 
	{A straightforward way to obtain strict equivariance for any backbone is group lifting and averaging~\cite{yarotsky2022universal,puny2022frame,basu2023equituning}, i.e., applying the network to all transformed copies of the input and averaging the inversely transformed outputs. This architecture-agnostic symmetrization~\cite{kim2023probabilistic}, however, acts only at the output: the backbone itself is not equivariant, and the transformed copies are processed independently without any interaction, which limits accuracy at comparable computational cost. As shown in Sec.~IV-F, it is not a competitive alternative to built-in equivariance for VMamba.}
	
	
	Recent works~\cite{cohen2016group,weiler2019general,shen2020pdoeconvs,2022Fconv,2025Bconv,hoogeboom2018hexaconv,weiler2018learning,shen2021pdo} have focused more on the principled  design of rotation equivariant neural networks that explicitly embed rotation equivariance into the architecture itself.
	For example, G-CNN~\cite{cohen2016group} was the first to successfully establish the group-equivariant convolutional framework, achieving \text{$\pi / 2$} rotation equivariance with rigorous theoretical guarantees.
	Building on group convolution, SFCNNs~\cite{weiler2018learning} and E2-CNN~\cite{weiler2019general} utilized the filter parametrization techniques for arbitrary-degree equivariance in the continuous domain.
	Subsequently, PDO-eConvs~\cite{shen2020pdoeconvs} and PDO-eS2CNNs~\cite{shen2021pdo} introduced the first equivariance error analysis for group equivariant convolutions, providing rigorous theoretical underpinnings.
	Furthermore, F-Conv~\cite{2022Fconv} leveraged Fourier series expansion to construct  high-accuracy filter-parameterized equivariant convolutions, achieving evidently better performance than classical non-equivariant CNNs on image super-resolution tasks.
	
	Beyond convolutional architectures, B-Conv~\cite{2025Bconv} introduced a rotation equivariant linear layer for multilayer perceptrons (MLPs) and constructed a rotation equivariant implicit neural representation framework for arbitrary-scale image super-resolution.
	For Transformer-based architectures, several studies~\cite{romero2020group,he2021efficient,xu20232,kundu2024steerable,hutchinson2021lietransformer,2025Bconv} have revealed that the core self-attention mechanism in ViTs possesses inherent rotational properties, leading to the design of equivariant ViTs.
	
	Despite these advances across CNNs and ViTs, research on rotation equivariant Mamba-based architectures remains largely unexplored.
	{The equivariance constraint imposed by a selective state space model is structurally different from the one imposed by convolution or self-attention.
	Convolution aggregates a neighborhood by a commutative summation and carries no state across positions, while self-attention is already permutation equivariant over tokens;
	in both cases equivariance is enforced by constraining the learnable components.
	Unlike these architectures, the selective SSM in VMamba operates fundamentally differently, as it aggregates the sequence along a fixed, non-learnable scanning trajectory, and its transition matrices are input-dependent\footnote{{See Sec.~S2.1 of the supplementary material for the detailed commonalities and distinctive characteristics of equivariant visual Mamba compared with equivariant CNNs and ViTs.}}.
	}

	The recent Spectral VMamba~\cite{2025spectralvmamba} represents an initial attempt, employing a rotation invariant scanning strategy derived from spectral decomposition.
	However, due to the absence of a rotational group dimension, this approach achieves only rotation \textit{invariance} rather than \textit{equivariance}, thereby discarding essential 
	directional information of local features and limiting its applicability to mid-level and low-level vision tasks.
	Moreover, the reliance on spectral decomposition introduces considerable computational overhead.
	In contrast, our proposed EQ-VMamba is a group-based equivariant architecture that preserves the orientation information of local features.
	Moreover, the proposed EQ-cross-scan strategy is simple and efficient, ensuring strict symmetry without incurring significant additional computational cost.

	\section{Rotation Equivariant Visual Mamba}\label{sec:Method}
	In this section, we present a principled framework for designing EQ-VMamba, a Mamba-based architecture with strict rotation equivariance under the $\mathrm{p}4$ group (i.e., $\{0^\circ, 90^\circ, 180^\circ, 270^\circ\}$)
	\footnote{{
	Beyond the $\mathrm{p}4$ group, our formulation can be applied to arbitrary finite cyclic transformation groups acting on the pixel grid by permutations. 
	The generalized formulation is given in Sec.~S1.5 of the supplementary material, and a reflection-equivariant instantiation is provided in Sec.~S3.1.	
	}}.
	Specifically,  we first establish the necessary preliminaries regarding the formulation of the Mamba architecture and the formal definition of rotation equivariance.
	Second, we provide a detailed exposition of the EQ-VMamba architecture, including: 
	1) the overall architecture; 
	2) the design of equivariant modules, including the EQ-patch embedding, the EQ-cross-scan/merge strategy, and EQ-VSS blocks; 
	3) a rigorous theoretical equivariance analysis for both the proposed individual modules and the entire network.
	Finally, we introduce two representative Mamba-based instantiations, EQ-VMamba and EQ-MambaIR, to demonstrate the framework's versatility across high-level, mid-level, and low-level vision tasks.

	\subsection{Preliminaries}\label{sec:Preliminaries}
	
	\textbf{Formulation of Mamba.}
	Mamba~\cite{gu2024mamba} extends structured state space sequence models (S4)~\cite{gu2021efficiently} by introducing a selective mechanism into the recurrent state propagation. 
	Specifically, Mamba maps an input sequence $\bm{x} \! \in \! \mathbb{R}^{L}$\footnote{For a multi-channel input sequence $\bm{x} \! \in \! \mathbb{R}^{L   \times   C}$, Mamba processes each channel independently and in parallel.} to an output sequence $\bm{y} \! \in \! \mathbb{R}^{L}$
	by maintaining a hidden state ${{\bm h}_i} \!\! \in \!\! \mathbb{R}^{N}$, which is updated via an input-dependent state transition matrix $ \bm{A}_i \! \in \! \mathbb{R}^{N \! \times \! N}$ as follows:
	\begin{equation}\label{eq:mamba}
		\begin{array}{l}
			\begin{aligned}
				{\bm{h}}_i &= {\bm A}_i {{\bm h}_{i-1}} + {{\bm B}_i} {{x}_i},\\
				{{ y}_i} &= {{\bm C}_i^{\top}} {{\bm h}_i} + {D} {{x}_i},
			\end{aligned}
		\end{array}
	\end{equation}
	where $\bm{A}_i$ is a diagonal matrix with elements bounded in $(0, 1]$,  ${\bm B}_i, {\bm C}_i \! \in \! \mathbb{R}^{N}$ are input-dependent parameters generated by linear projections, and $D \! \in \! \mathbb{R}$ denotes a skip-connection coefficient.
	Here, $L$ and $N$ represent the sequence length and hidden state dimension, respectively.
	The selective nature of $\bm{A}_i$ enables the model to dynamically propagate or forget historical information, significantly enhancing its capability for long-range dependency modeling.

	\textbf{Definition of Rotation Equivariance.}
	Rotation equivariance requires that applying a rotation transformation to the input induces a predictable, corresponding transformation of the output, thereby preserving the structural relationship between features. 
	Formally, let $\Psi$ be a mapping from an input space to an output space, and $\mathcal G$ be a group of rotation transformations defined as $\mathcal{G}\!=\l\{  {G}_t | t = 1,2,\cdots, T\r\}$, where
	\begin{equation}\label{eq:Group}
		{G}_t =
		\begin{bmatrix}
			\cos \nicefrac{2\pi (t-1)}{T} &  \sin \nicefrac{2\pi  (t-1)}{T} \\
			-\sin \nicefrac{2\pi  (t-1)}{T} &  \cos \nicefrac{2\pi  (t-1)}{T}
		\end{bmatrix} .
	\end{equation}
	The mapping $\Psi(\cdot)$ is equivariant with respect to ${\mathcal G}$ if, for any rotation matrix ${ G} \in {\mathcal G}$, it satisfies:
	\begin{equation}\label{eq:equivariance}
		\Psi \( \pi^{\scriptscriptstyle {\mathcal R}}_{\scriptscriptstyle { G}} \({\bm I}\)\)  = {\tilde{\pi}}^{\scriptscriptstyle {\mathcal R}}_{\scriptscriptstyle { G}} \(\Psi \({\bm I}\)\),
	\end{equation}
	where $\pi^{\scriptscriptstyle {\mathcal R}}_{\scriptscriptstyle G} (\cdot)$ denotes the spatial rotation transformation $ G$ acting on the input $\bm I$,  and ${\tilde{\pi}}^{\scriptscriptstyle {\mathcal R}}_{\scriptscriptstyle G} (\cdot)$ denotes the induced transformation on the output $\Psi({\bm I})$.
	{Note that rotation invariance is an important special case of rotation equivariance, which arises when the induced output transformation degenerates to the identity, i.e., $\tilde{\pi}^R_{G}=\mathrm{Id}$, in which case $\Psi(\pi^R_G(I))=\Psi(I)$.}

	\begin{figure}[t]
		\centering
		\includegraphics[width=0.48\textwidth]{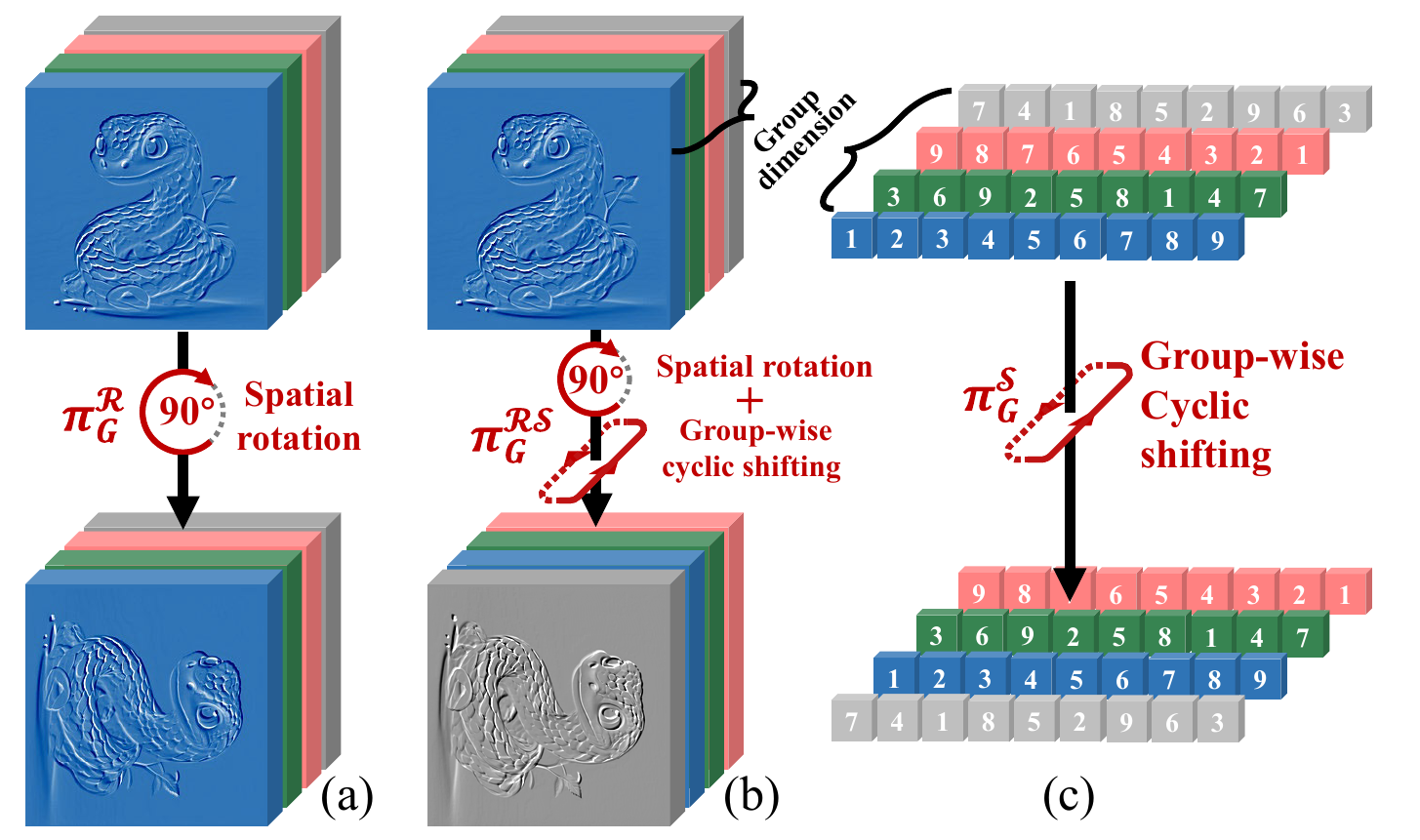}
		\vspace{-1mm}
		\caption{
			Illustration of the group dimension (indexed by colors),  and three fundamental transformation operators utilized in our equivariant framework. (a)-(c) illustrate $\pi^{\scriptscriptstyle {\mathcal R}}_{\scriptscriptstyle G} (\cdot)$, $\pi^{\scriptscriptstyle {\mathcal RS}}_{\scriptscriptstyle G} (\cdot)$, and $\pi^{\scriptscriptstyle {\mathcal S}}_{\scriptscriptstyle G} (\cdot)$, respectively.
		}
		\label{fig:notations_transformation}
		\vspace{-4mm}
	\end{figure}
	
	\textbf{Group Dimension and Transformations.}
	As illustrated in Fig.~\ref{fig:notations_transformation}, in group-based equivariant networks, the feature map typically incorporates an additional dimension, namely the group dimension{, a representation introduced by G-CNN~\cite{cohen2016group} and standard in subsequent group-equivariant architectures}.
	Specifically, whereas feature maps in conventional networks have the shape $H\times W\times \hat{C}$ (where $H$, $W$, and $\hat{C}$ denote the height, width, and number of channels, respectively), those in group-based equivariant networks are of shape $H\times W\times C\times T$, where $T=|\mathcal{G}|$ is the number of elements in the group and $C$ is typically set to $\frac{\hat{C}}{T}$ to maintain computational parity.
	As will become evident in the subsequent sections, this additional group dimension plays a crucial role in preserving directional information and enabling the construction of equivariant networks.
	Following the notation of~\cite{shen2020pdoeconvs}, 
	for a feature map $\bm{X}\in \mathbb{R}^{H\times W\times C\times T}$, we denote its group component corresponding to $G\in \mathcal{G}$ as $\bm{X}^G \in \mathbb{R}^{H\times W\times C}$, where $G$ serves both as a rotation matrix and as an index along the group dimension.

	\begin{figure*}[t]
		\centering
		\includegraphics[width=0.99\textwidth]{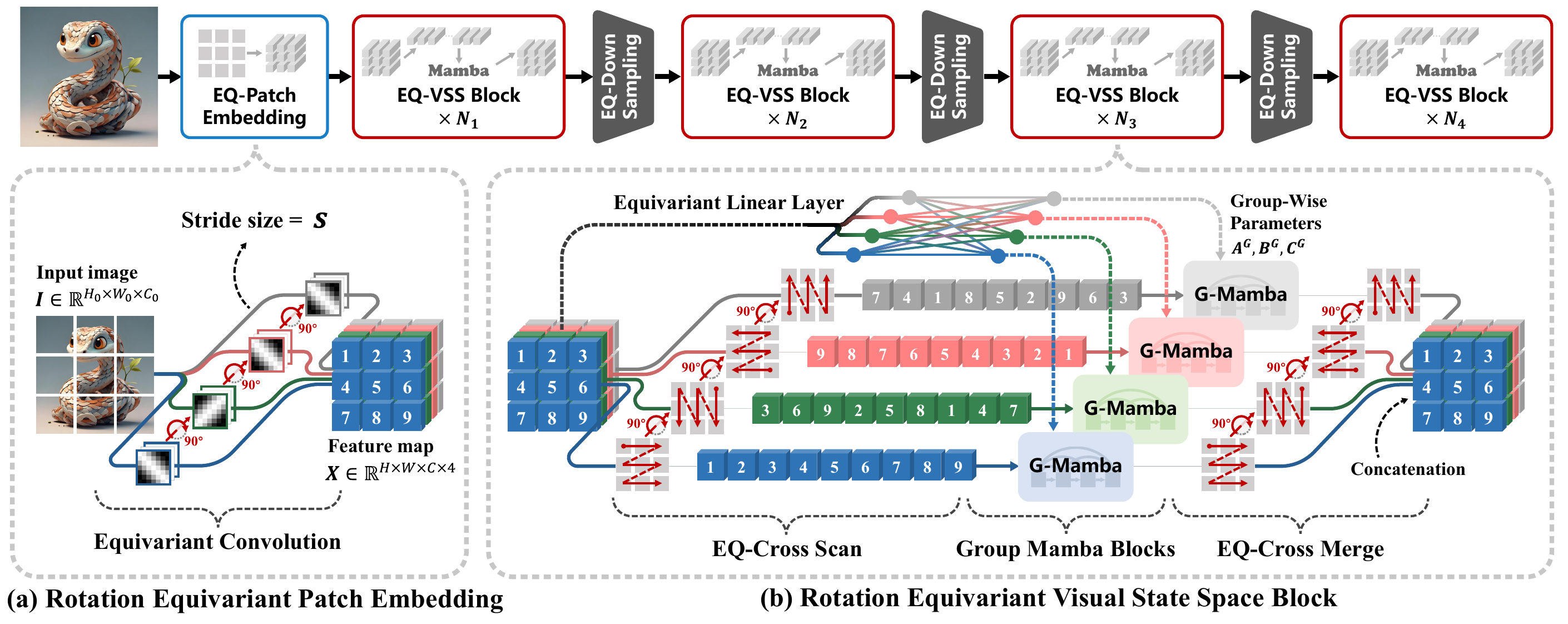}
		\vspace{-3mm}
		\caption{
			Overall architecture of the proposed end-to-end rotation equivariant visual Mamba (EQ-VMamba). 
			The framework mainly comprises: (a) an EQ-patch embedding module that tokenizes the input image into group-structured feature maps,
			and (b) a stack of EQ-VSS blocks for hierarchical feature extraction.
			Each EQ-VSS block integrates an EQ-cross-scan operation for image-to-sequence flattening, group Mamba blocks for sequence modeling, and an EQ-cross-merge operation for sequence-to-image reconstruction.
		}
		\label{fig:EQ-VMamba_architecture}
		\vspace{-4mm}
	\end{figure*}

	While rotation transformations on images are intuitively straightforward, on feature maps equipped with a group dimension, spatial rotation becomes coupled with a cyclic shifting along the group dimension. Specifically, in this paper, our framework primarily involves three types of transformations: 
	
	(I) $\pi^{\scriptscriptstyle {\mathcal R}}_{\scriptscriptstyle G} (\cdot)$: rotation in the spatial domain without altering the group dimension. As shown in Fig.~\ref{fig:notations_transformation}(a), this transformation is typically applied to input images. 
	
	(II) $\pi^{\scriptscriptstyle {\mathcal RS}}_{\scriptscriptstyle G} (\cdot)$: rotation in the spatial domain combined with a cyclic shifting along the group dimension, as shown in Fig.~\ref{fig:notations_transformation}(b). Formally, for any feature map $\bm{X}\in\mathbb{R}^{H\times W\times C\times T}$, we have: 
	\begin{equation}\label{eq:piRS}
		\l(\pi^{\scriptscriptstyle {\mathcal RS}}_{\scriptscriptstyle \hat{G}} (\bm{X})\r)^G =\pi^{\scriptscriptstyle {\mathcal R}}_{\scriptscriptstyle \hat{G}} \l(\bm{X}^{\hat{G}^{-1}G}\r), ~~ \forall G, \hat{G} \in \mathcal{G}.
	\end{equation}
	This transformation is particularly common in prior equivariant CNNs and is typically applied to feature maps that carry a group dimension. 
	
	(III) $\pi^{\scriptscriptstyle {\mathcal S}}_{\scriptscriptstyle G} (\cdot)$:  cyclic shifting along the group dimension, with no change in the spatial domain. As shown in Fig.~\ref{fig:notations_transformation}(c), for a feature map $\bm X$, we have 
	\begin{equation}\label{eq:piS}
		(\pi^{\scriptscriptstyle {\mathcal S}}_{\scriptscriptstyle \hat{G}} (\bm{X}))^G = \bm{X}^{\hat{G}^{-1}G}, ~~ \forall G, \hat{G}\in \mathcal{G}.
	\end{equation}
	This transformation is applied to the scanning order of Mamba in our proposed framework.
	
	Throughout this paper, we focus on the design of EQ-VMamba under the 90-degree rotation group (i.e., the $\mathrm{p}4$ group), corresponding to $T=4$.

	\subsection{Overall Architecture of EQ-VMamba}
	{To construct an end-to-end rotation equivariant visual Mamba architecture, we design each constituent module to strictly satisfy the equivariance constraint defined in Eq.~\eqref{eq:equivariance}, so that equivariance is preserved under composition.}
	Driven by this requirement, we adopt VMamba~\cite{2024Vmamba}---one of the most representative visual Mamba backbones---as our baseline, and systematically reformulate all non-equivariant components into their rotation equivariant counterparts, yielding the proposed EQ-VMamba architecture.
	The overall architecture of the EQ-VMamba backbone is illustrated in Fig.~\ref{fig:EQ-VMamba_architecture}.
	Specifically, it consists of the following key equivariant modules.

	\textbf{Rotation Equivariant Patch Embedding.}
	The vanilla VMamba begins with a patch embedding layer that partitions the input image into patches and tokenizes them into an initial feature map.
	However, this standard patch embedding module lacks rotation equivariance.
	To address this, we develop a rotation equivariant patch embedding that explicitly encodes the orientation information of each patch into the rotation group dimension of the feature map.
	
	\begin{figure}[t]
		\centering
		\includegraphics[width=0.5\textwidth]{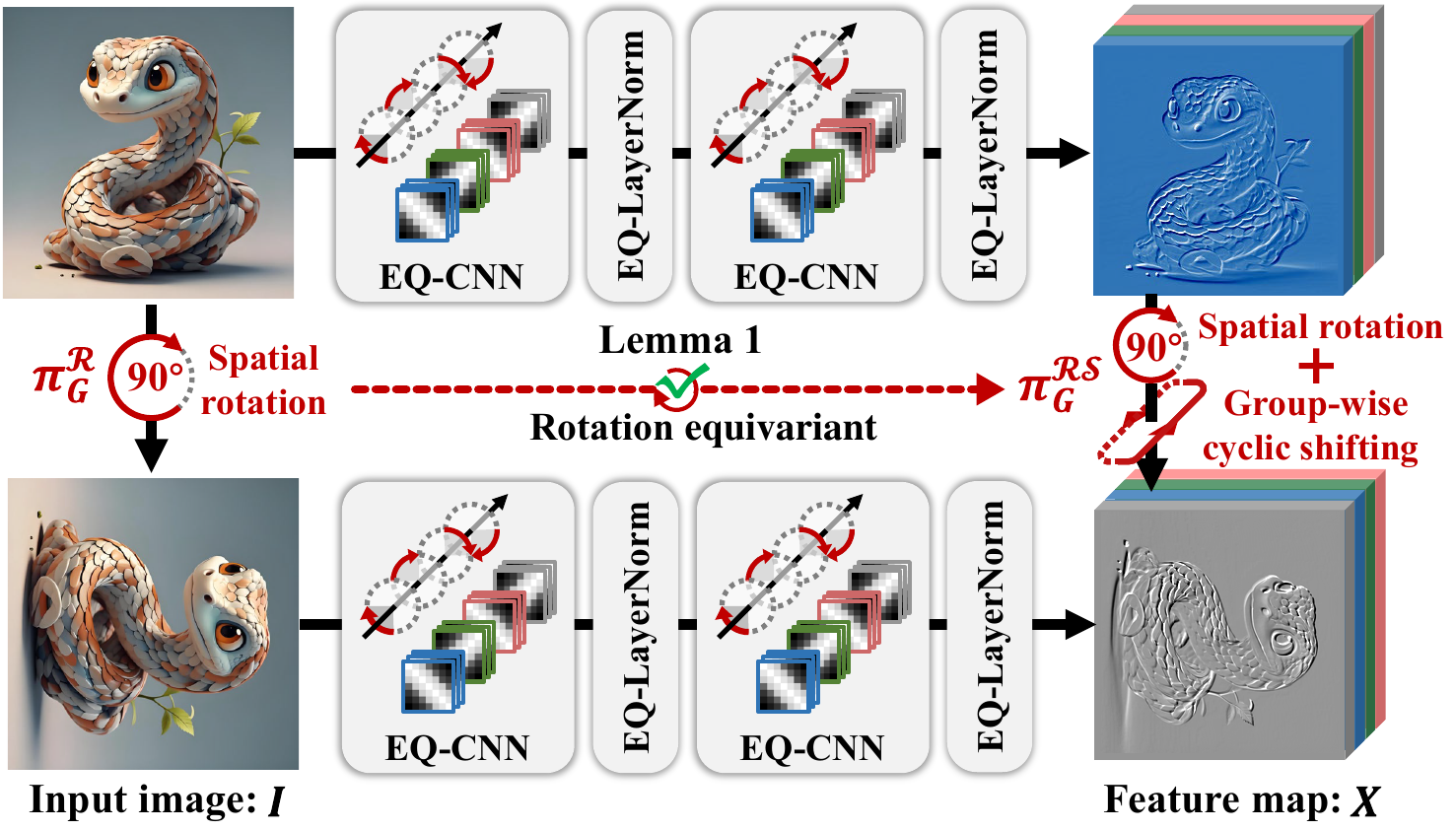}
		\vspace{-6.5mm}
		\caption{
			Illustration of the rotation equivariant patch embedding.
			A spatial rotation applied to the input image induces a joint transformation on the output feature map, i.e., $\mathrm{PE}_{eq} \( {{\pi}}^{\scriptscriptstyle {\mathcal{R}}}_{\scriptscriptstyle G}  \({\bm I}  \) \)  = {{\pi}}^{\scriptscriptstyle {\mathcal{RS}}}_{\scriptscriptstyle G} \( \mathrm{PE}_{eq}  \({\bm I} \)\)$.
		}
		\label{fig:EQ_CNN}
		\vspace{-5mm}
	\end{figure}
	
	\textbf{Rotation Equivariant Visual State-Space Block.}
	Following patch embedding, VMamba utilizes a stack of multiple Visual State-Space (VSS) blocks for hierarchical feature extraction.
	Each standard VSS block consists of three stages: 
	1) flattening the 2D feature map into four 1D sequences via the cross-scan mechanism;
	2) processing these sequences with four independent Mamba blocks;
	and 3) reconstructing the 2D feature map from output sequences via cross-merge (i.e., the inverse operation of cross-scan).
	However, both the cross-scan/merge operations and the independently parameterized Mamba blocks lack rotation equivariance.
	To address this, we first propose rotation equivariant cross-scan/merge to ensure equivariance during the flattening and reconstruction stages.
	Furthermore, we introduce the group Mamba blocks to ensure rotation equivariance during the sequence modeling stage.
	By integrating these components, we construct a rotation equivariant EQ-VSS block.	
	
	Beyond the patch embedding and EQ-VSS blocks, other essential components within VMamba---such as depthwise convolution and down/up-sampling---also require equivariant modifications to ensure rotation equivariance throughout the network. 
	{Rotation equivariant counterparts of these standard modules have already been established in previous studies~\cite{weiler2019general,shen2020pdoeconvs,2022Fconv}, allowing us to construct a fully end-to-end EQ-VMamba framework.}
	By replacing all non-equivariant components with their equivariant counterparts, we obtain an end-to-end rotation equivariant Mamba-based architecture that maintains the original Mamba's efficiency while significantly enhancing its robustness to rotation transformations.

	\subsection{Design of Rotation Equivariant Patch Embedding}
	The vanilla patch embedding in VMamba performs patch embedding using overlapping strided convolutional layers followed by LayerNorm2d.
	However, this standard patch embedding is not rotation equivariant and fails to preserve the orientation information of input image patches.
	To overcome this limitation, we develop a rotation equivariant patch embedding (EQ-patch embedding).
	{Specifically, we leverage established rotation equivariant CNN (EQ-CNN) layers~\cite{cohen2016group,weiler2018learning,shen2020pdoeconvs,2022Fconv} to tokenize the input image into the group-structured feature map required by the subsequent equivariant scanning and recurrence in Sec.~III-D.}
	As illustrated in Fig.~\ref{fig:EQ_CNN}, the EQ-CNN layer $\Psi (\cdot)$ achieves rotation equivariance by sharing a base kernel ${\bm W}_{\Psi}$ across discrete rotations within a defined group.
	Formally, given an input image ${\bm I} \! \in \! \mathbb{R}^{H_0 \! \times  \! W_0 \! \times  \! C_0}$, the EQ-CNN layer $\Psi (\cdot)$ maps it to a group-structured feature map ${\bm \hat{X}} \! \in  \! \mathbb{R}^{\frac{H_0}{s}\!\times \! \frac{W_0}{s} \! \times \!   C \! \times  \! 4}$ as follows:
	\begin{equation}\label{eq:EQCNN}
		\begin{aligned}
			{\bm \hat{X}} = \Psi \( {\bm I}\), ~\mbox{with}~ {\bm \hat{X}}^{G} = \pi^{\scriptscriptstyle{\mathcal R}}_{\! \scriptscriptstyle G} \! \l( 	\bm W_{\Psi} \r)  * {\bm I} + {\bm b_{\Psi}}, ~ \forall G\in \mathcal{G},
		\end{aligned}
	\end{equation}
	where $*$ denotes the convolution operation, $s$ is the stride step, $C$ is the output channel number, and
	${\bm W_{\Psi}} \! \in \! \mathbb{R}^{K\!\times \! K \! \times \!  C_0 \! \times \! C}$ and $\bm{b}_{\Psi} \! \in \! \mathbb{R}^{C}$ denote the learnable weights and biases, respectively.

	\begin{figure*}[t]
		\centering
		\includegraphics[width=0.99\textwidth]{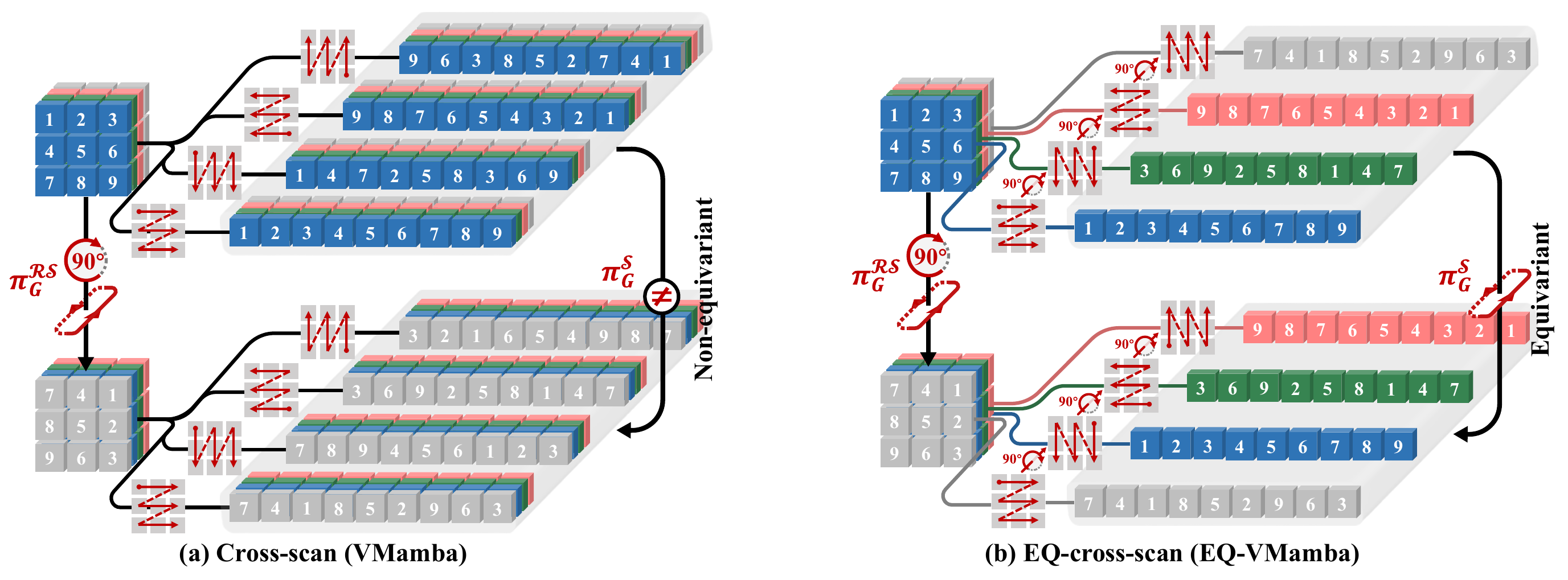}
		\vspace{-3mm}
		\caption{
			Comparison between the non-equivariant cross-scan and the proposed equivariant EQ-cross-scan.
			(a) Under a 90$^\circ$ rotation of the input image, the standard cross-scan in VMamba fails to maintain sequence consistency.
			(b) In contrast, EQ-cross-scan preserves the ordering of tokens within each 1D sequence, ensuring rigorous structural correspondence between the rotated and unrotated representations.
		}
		\label{fig:EQ-Cross-Scan}
		\vspace{-4mm}
	\end{figure*}

	By stacking  multiple EQ-CNN layers (typically two layers with a stride of 2) and EQ-LayerNorm layers~\cite{weiler2019general}, we construct an equivariant EQ-patch embedding block $\mathrm{PE}_{eq}(\cdot)$ that maps an input image to a group-structured feature map:
	\begin{equation}\label{eq:EQ-patch embedding}
		\begin{aligned}
			{\bm X} = \mathrm{PE}_{eq}(\bm I).
		\end{aligned}
	\end{equation}
	{For this structural formulation, prior studies on equivariant CNNs~\cite{cohen2016group,2022Fconv} establish the following lemma:}
	\begin{Lemma}\label{lemma:EQ_CNN}
		Let ${\bm I} \in \mathbb{R}^{H_0 \times W_0 \times C_0}$ be an input image and let $\mathrm{PE}_{eq}(\cdot)$ denote the EQ-patch embedding defined in ~\eqref{eq:EQ-patch embedding}. 
		For any group element ${G} \in {\mathcal G}$, the following equivariance property holds:
		\begin{equation}\label{eq:EQ-Cross-Scan_Equivariance}
			\mathrm{PE}_{eq} \( {{\pi}}^{\scriptscriptstyle {\mathcal{R}}}_{\scriptscriptstyle G}  \({\bm I}  \) \)  = {{\pi}}^{\scriptscriptstyle {\mathcal{RS}}}_{\scriptscriptstyle G} \( \mathrm{PE}_{eq}  \({\bm I} \)\).
		\end{equation}
	\end{Lemma}
	Lemma~\ref{lemma:EQ_CNN} indicates that a spatial rotation applied to the input image induces a corresponding spatial rotation together with a cyclic shifting across the rotation group dimension in the output feature map.

	\subsection{Design of Rotation Equivariant Visual State-Space Block}
	The core contribution of EQ-VMamba is the design of a rotation equivariant Visual State-Space (EQ-VSS) block, which primarily incorporates a novel rotation equivariant cross-scan/merge strategy and group Mamba blocks.
	
	\textbf{Rotation Equivariant Cross-Scan/Merge.}
	As formulated in Eq.~\eqref{eq:mamba}, Mamba is inherently designed for processing one-dimensional sequential data.
	To apply this mechanism to 2D image data, VMamba employs a four-way cross-scan strategy that flattens 2D image tokens into four distinct 1D sequences~\cite{2024Vmamba}.
	However, this vanilla cross-scan strategy fails to satisfy rotation equivariance.
	As illustrated in Fig.~\ref{fig:EQ-Cross-Scan} (a), rotating the input image induces inconsistent and misaligned transformations on the resulting sequences.

	To achieve rigorous equivariance, we propose a simple yet effective rotation equivariant cross-scan (EQ-cross-scan) strategy for image-to-sequence flattening.
	As illustrated in Fig.~\ref{fig:EQ-Cross-Scan} (b), EQ-cross-scan employs four rotationally symmetric scanning paths, each independently processing one component of the feature map along the group dimension.
	Formally, given a group-structured 2D feature map ${\bm X} \! \in  \! \mathbb{R}^{H\!\times \! W \! \times \!   C \! \times  \! 4}$, we define:
	\begin{equation}\label{eq:EQ-Cross-Scan}
		\begin{aligned}
			{\bm x} = \tau_{eq}\l(\bm X \r), ~\mbox{with}~ {\bm x}^{G} = \tau \l( \pi_{\scriptscriptstyle G^{-1}}^{\scriptscriptstyle{\mathcal R}}\l( {\bm X}^{G} \r)\r), ~ \forall  G\in \mathcal{G},
		\end{aligned}
	\end{equation}
	where ${\bm x} \! \in  \! \mathbb{R}^{H \! W \! \times \! C \! \times \! 4}$ denotes the resulting  1D feature sequences, and $\tau (\cdot)$ represents the base scanning path (corresponding to the standard matrix-to-vector unfolding).
	
	Symmetrically, we define the inverse operation of $\tau_{eq} (\cdot)$, termed EQ-cross-merge $\tau_{eq}^{inv} (\cdot)$ for equivariant sequence-to-image reconstruction:
	\begin{equation}\label{eq:EQ-Cross-Merge}
		\begin{aligned}
			{\bm \hat{X}} \! = \tau_{eq}^{inv} \l( {\bm x} \r)\!, 
			~\mbox{with}~
			{\bm \hat{X}}^{G} \!= \pi_{\scriptscriptstyle G}^{\scriptscriptstyle{\mathcal R}} \l(\tau^{inv} \l(  {\bm x}^{G} \r)\r)\!, ~ \forall G\in \mathcal{G},
		\end{aligned}
	\end{equation}
	where $\tau^{inv} (\cdot)$ denotes the standard vector-to-matrix folding. 
	
	Based on this formulation, we establish the following theoretical guarantees (detailed proofs are provided in the supplementary material).

	\begin{Theorem}\label{theorem:EQ_cross_scan}
		Let ${\bm X} \in \mathbb{R}^{H \times W \times C \times 4}$ be a group-structured feature map and let  ${\bm x} \in \mathbb{R}^{HW \times C \times 4}$ be a group-structured feature sequence. 
		Let $\tau_{eq}(\cdot)$ and $\tau_{eq}^{inv}(\cdot)$ denote the EQ-cross-scan and EQ-cross-merge operators defined in \eqref{eq:EQ-Cross-Scan} and \eqref{eq:EQ-Cross-Merge}, respectively. 
		For any group element ${\hat G} \in {\mathcal G}$, the following equivariance properties hold:
		\begin{equation}\label{eq:EQ-Cross-Scan_Equivariance2}
			\tau_{eq} \( {{\pi}}^{\scriptscriptstyle {\mathcal{RS}}}_{\scriptscriptstyle {\hat G}}  \({\bm X}  \) \)  = {{\pi}}^{\scriptscriptstyle {\mathcal{S}}}_{\scriptscriptstyle {\hat G}} \( \tau_{eq}  \({\bm X} \)\),
		\end{equation}
		\begin{equation}\label{eq:EQ-Cross-Merge_Equivariance}
			\tau_{eq}^{inv} \( {{\pi}}^{\scriptscriptstyle {\mathcal{S}}}_{\scriptscriptstyle {\hat G}}  \({\bm x}\) \) = {{\pi}}^{\scriptscriptstyle {\mathcal{RS}}}_{\scriptscriptstyle {\hat G}} \( \tau_{eq}^{inv} \({\bm x}\)\).
		\end{equation}
	\end{Theorem}
	
	Theorem~\ref{theorem:EQ_cross_scan} establishes that both the EQ-cross-scan and EQ-cross-merge  are strictly equivariant under discrete 90-degree rotations.
	An intuitive illustration is provided in Fig.~\ref{fig:EQ-Cross-Scan} (b): when the input image undergoes a 90$^\circ$ spatial rotation and a group-wise cyclic shifting, the resulting 1D sequences generated by EQ-cross-scan differ only by the corresponding permutation along the group dimension, ensuring the equivariance of the image-to-sequence flattening process.
	
	{ It is worth noting that while Fig.~\ref{fig:EQ-Cross-Scan} illustrates the standard EQ-cross-scan, our theoretical formulation does not rely on a specific scanning trajectory. By treating the base scanning path $\tau(\cdot)$ in Eq.~\eqref{eq:EQ-Cross-Scan} as an arbitrary mapping, our framework effectively converts any deterministic scanning operation into an equivariant counterpart\footnote{{Further details and experimental results under diverse scanning trajectories are provided in Sec.~S1.6 and S3.2 of the supplementary material.}}.}

	\textbf{Group Mamba Blocks.}
	Following the image-to-sequence flattening process, the vanilla VMamba employs four independent Mamba blocks to process each of the four feature sequences in parallel. 
	However, under rotation transformations, the group-dimensional components of feature maps will be processed by different Mamba blocks compared to the unrotated case, producing a non-equivariant output.
	Consequently, the independent parameterization of the Mamba blocks in VMamba lacks rotation equivariance.
	
\begin{figure*}[t]
		\centering
		\includegraphics[width=0.95\textwidth]{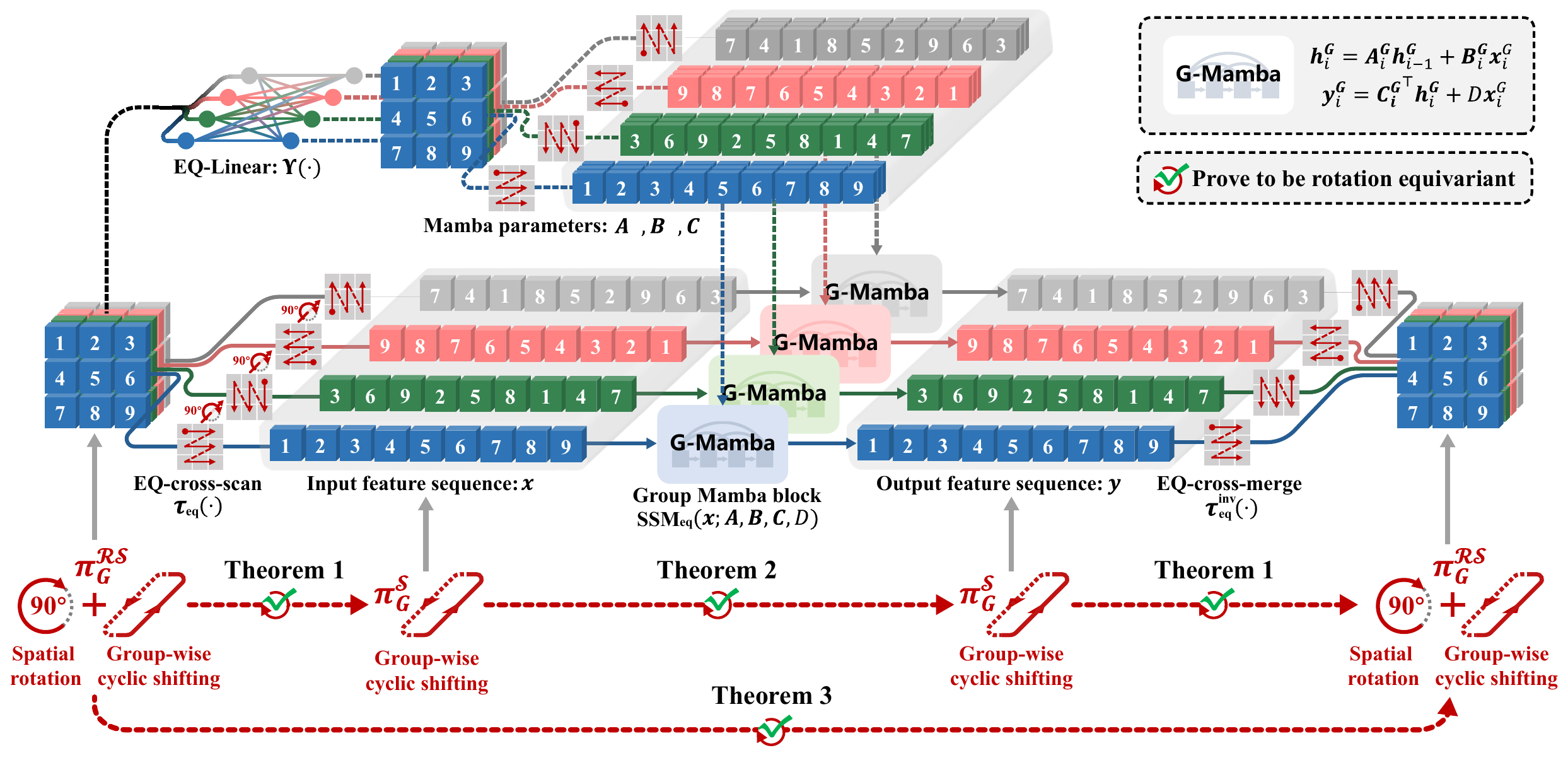}
		\vspace{-3mm}
		\caption{\textbf{Upper}: Architectural pipeline of the proposed equivariant Visual State-Space (EQ-VSS) block.
			(1) The block first employs EQ-Linear layers to generate input-dependent Mamba parameters ${\bm A}$, ${\bm B}$, and ${\bm C}$.
			(2) These parameters, along with the input feature map, are partitioned along the group dimension and flattened into four 1D sequences via EQ-cross-scan.
			(3) Each feature sequence is processed in parallel by the group Mamba block using its corresponding group-wise parameters.
			(4) Finally, the output feature map is restored through the EQ-cross-merge operation.
			\textbf{Bottom}: Illustration of the proposed theoretical results, which show the correspondence between each theorem and its respective
			network module.
		}
		\label{fig:EQ-Mamba_Blocks}
		\vspace{-4mm}
	\end{figure*}

	To address this issue, we construct the group Mamba blocks by restructuring the learnable parameters ${\bm A}, {\bm B}, $ and ${\bm C}$ in the Mamba blocks.
	Specifically, as illustrated in Fig.~\ref{fig:EQ-Mamba_Blocks}, we first replace the non-equivariant linear layers with EQ-Linear layers $\Upsilon (\cdot)$, while retaining all element-wise non-linear operations from the original Mamba formulation.
	Furthermore, to fully exploit feature interactions along the group dimension, the parameters ${\bm A}$, ${\bm B}$, and ${\bm C}$ are generated prior to the scanning operation and subsequently  flattened into one-dimensional sequences consistent with the input sequence $\bm x$.
	This design enables an equivariant reformulation while preserving the original parameter generation scheme of Mamba.
	We illustrate the construction using parameter ${\bm A}$ as an example:
	\begin{equation}\label{eq:EQ-Mamba_Blocks_A}
		\begin{aligned}
			{\bm {A}} = \tau_{eq} \( \delta_{a} \(\Upsilon_{a} \(\bm X \)\) \)        
		\end{aligned}
	\end{equation}
	where ${\bm X} \! \in  \! \mathbb{R}^{H \! \times \! W \! \times \! C \! \times \! 4}$, ${\bm A} \! \in  \! \mathbb{R}^{H W \! \times \! N \! \times \!  N \! \times \! C \! \times \! 4 }$, 
	$\delta_{a} \( \cdot \)$ includes the element-wise non-linear operation and the reshape/diagonal operations ($\mathbb{R}^{H \! \times \! W \! \times \! N  \! C \! \times \! 4 } \to \mathbb{R}^{H \! \times \! W \! \times \! N \! \times \!  N \! \times \! C \! \times \! 4 }$) from the original Mamba framework,
	$\tau_{eq} \( \cdot \)$ denotes the EQ-cross-scan operator defined in Eq.~\eqref{eq:EQ-Cross-Scan}, 
	and $ \Upsilon_{a} \l(\cdot \r)$ denotes an EQ-Linear layer~\cite{ravanbakhsh2020universal,2025Bconv} defined as
	\begin{equation}\label{eq:EQLinear}
		\begin{aligned}
			{\l( \Upsilon_{a} \l(\bm X \r) \r)}^{G} = {\bm X}  \times  {{\pi}}^{\scriptscriptstyle \mathcal{S}}_{\! \scriptscriptstyle G} \! \l({\bm W}_{a}\r)
			+ {\bm b_{a}}, ~ \forall G \in \mathcal{G}
		\end{aligned}
	\end{equation}
	where ${\bm W}_{a}\! \in  \! \mathbb{R}^{4C \! \times \! N \! C}$ and $\bm b_{a} \! \in  \! \mathbb{R}^{N \! C}$ are learnable parameters.

	Similarly, parameters ${\bm B}$ and ${\bm C}$ can be generated by 
	\begin{equation}\label{eq:EQ-Mamba_Blocks_BC}
		\begin{aligned}
			{\bm B} = \tau_{eq} \l(  \delta_{b} \( \Upsilon_{b} \l(\bm X \r) \) \r), ~
			{\bm C} = \tau_{eq} \l( \delta_{c} \( \Upsilon_{c} \l(\bm X \r)  \) \r), 
		\end{aligned}
	\end{equation}
	where
	${\bm B}, {\bm C} \! \in  \! \mathbb{R}^{H W \! \times \!   N \! \times  \! 4}$,   
	$\delta_{b} \( \cdot \)$ and $\delta_{c} \( \cdot \)$ denote the reshape operation,
	and $ \Upsilon_{b} \l(\cdot \r)$ and  $\Upsilon_{c} \l(\cdot \r)$ denote EQ-Linear layers.
	Then, parameters ${\bm A}, {\bm B}, {\bm C}$, together with a shared scalar parameter ${D}$, are assigned to the group Mamba blocks for sequence modeling:
	\begin{equation}\label{eq:group_mamba}
		\begin{array}{l}
			\begin{aligned}
				{\bm y} = \mathrm{SSM_{eq}} \( {\bm x}; {\bm A}, {\bm B}, {\bm C}, D \),
			\end{aligned}
		\end{array}
	\end{equation}
	where ${\bm y} \! \in  \! \mathbb{R}^{H W \! \times \!   C \! \times  \! 4}  $ denotes the output sequences.
	The group Mamba blocks comprise four parallel Mamba blocks, each modeling its corresponding group-wise feature sequences as
	\begin{equation}\label{eq:each_group_mamba}
		\begin{array}{l}
			\begin{aligned}
				{\bm y}^{G} = \mathrm{SSM} \( {\bm x}^{G}; {\bm A}^{G}, {\bm B}^{G}, {\bm C}^{G}, D \), ~\forall G \in \mathcal{G},
			\end{aligned}
		\end{array}
	\end{equation}
	where $\mathrm{SSM} \( \cdot \) $ denotes the standard Mamba formulation defined in Eq.~\eqref{eq:mamba}. Concretely, for $k=1,2,\cdots, C$ and $\forall G \in \mathcal{G}$,
	\begin{equation}\label{eq:group_mamba2}
		\begin{array}{l}
			\begin{aligned}
				{{\bm h}}^{G}_{i, k} &= {\bm A}^{G} _{i, k} {{\bm h}^{G}_{i-1, k}} + {{\bm B}^{G}_{i, k}} {{\bm x}^{G}_{i, k}},\\
				{{\bm y}^{G}_{i, k}} &= {{{\bm C}^{G}_{i, k}}^{\top}} {{\bm h}^{G}_{i, k}} + {D} {{\bm x}^{G}_{i, k}}.
			\end{aligned}
		\end{array}
	\end{equation}
	Finally, the output 2D feature map is recovered through EQ-cross-merge, i.e., ${\bm Y} = \tau_{eq}^{inv} ({\bm y})$.
	Based on this formulation, we establish the following theoretical guarantees.
	
	\begin{Theorem}\label{theorem:group_mamba}
		Let ${\bm X} \in \mathbb{R}^{H \! \times \! W  \! \times \! C \! \times \! 4}$ be a group-structured feature map.
		Given the group Mamba blocks ${\mathrm{SSM_{eq}}}(\cdot)$ defined in \eqref{eq:group_mamba} with parameters ${\bm A}  $, ${\bm B}$, and ${\bm C}$ derived according to \eqref{eq:EQ-Mamba_Blocks_A} and \eqref{eq:EQ-Mamba_Blocks_BC}, 
		the following equivariance property holds for any ${{\hat G}} \in {\mathcal G}$:
		\begin{equation}\label{eq:Group_Mamba_block_Equivariance}
			\begin{aligned}
				{\mathrm{SSM_{eq}}}& \( 
				{{\pi}}^{\scriptscriptstyle {\mathcal{S}}}_{\scriptscriptstyle {\hat G}} \( {\bm x} \); 
				{{\pi}}^{\scriptscriptstyle {\mathcal{S}}}_{\scriptscriptstyle {\hat G}} \( {\bm A} \), 
				{{\pi}}^{\scriptscriptstyle {\mathcal{S}}}_{\scriptscriptstyle {\hat G}} \( {\bm B} \),
				{{\pi}}^{\scriptscriptstyle {\mathcal{S}}}_{\scriptscriptstyle {\hat G}} \( {\bm C} \),
				D \)  \\
				& ~~~~~~~~ ~~~~~~~~ ~ = {{\pi}}^{\scriptscriptstyle {\mathcal{S}}}_{\scriptscriptstyle {\hat G}} \({\mathrm{SSM_{eq}}} \({\bm x}; {\bm A}, {\bm B}, {\bm C}, D \)\).
			\end{aligned}	
		\end{equation} 
	\end{Theorem}
	
	Theorem~\ref{theorem:group_mamba} establishes that the proposed group Mamba blocks strictly satisfy 90-degree rotation equivariance.

	\textbf{Entire Formulation of EQ-VSS Block.}
	By combining EQ-cross-scan (Eq.~\eqref{eq:EQ-Cross-Scan}), the group Mamba block (Eq.~\eqref{eq:group_mamba}), and EQ-cross-merge (Eq.~\eqref{eq:EQ-Cross-Merge}), we formulate the equivariant EQ-VSS block for an input feature map $\bm X \in \mathbb{R}^{H \times W \times C \times T}$ as
	\begin{equation}\label{eq:EQ-VSS_Block}
		{\mathrm{VSS_{eq}}}({\bm X}) = \tau_{eq}^{inv} \( {\mathrm{SSM_{eq}}} \( \tau_{eq} \( {\bm X}\); {\bm A},  {\bm B},  {\bm C}, D \)\).
	\end{equation}
	Building on Theorems~\ref{theorem:EQ_cross_scan} and \ref{theorem:group_mamba}, we establish the rotational equivariance of the entire EQ-VSS block as follows.
	
	\begin{Theorem}\label{theorem:EQ-VSS_block}
		Under the same conditions as Theorems~\ref{theorem:EQ_cross_scan} and \ref{theorem:group_mamba}, 
		the EQ-VSS block defined in Eq.~\eqref{eq:EQ-VSS_Block} satisfies the equivariance property for any ${{\hat G}} \in {\mathcal G}$:
		\begin{equation}\label{eq:EQ-Mamba_Blocks_rotate3}
			{\mathrm{VSS_{eq}}} \( \pi^{\scriptscriptstyle {\mathcal{RS}}}_{\scriptscriptstyle {\hat G}} \({\bm X} \) \)  
			= \pi^{\scriptscriptstyle {\mathcal{RS}}}_{\scriptscriptstyle {\hat G}}  \( {\mathrm{VSS_{eq}}} \( {\bm X} \) \).
		\end{equation}
	\end{Theorem}
	
	Fig.~\ref{fig:EQ-Mamba_Blocks} provides an intuitive illustration: a spatial rotation combined with a group-wise cyclic shifting applied to the input feature map results in identical transformations at the output of the EQ-VSS block, thereby empirically validating the theoretical equivariance property established in Theorem~\ref{theorem:EQ-VSS_block}.

	
	\textbf{Remarks.} 
	By combining Lemma~\ref{lemma:EQ_CNN} and Theorem~\ref{theorem:EQ-VSS_block}, the entire EQ-VMamba backbone achieves end-to-end equivariance under 90-degree rotations. {Theorems~1--3 establish the equivariance of the proposed construction rather than the uniqueness of this route to equivariance.}
	
	Moreover, it is worth emphasizing that, in the original VMamba architecture, the linear layer used to generate $\bm A$, $\bm B$, and $\bm C$ contains $\hat{C} \times N\hat{C}$ parameters, where $\hat{C}$ denotes the number of feature channels in the original VMamba network. To preserve consistency in the total number of feature channels, we set the channel dimension of EQ-VMamba, $C$, to $\hat{C}/4$. Under this setting, the number of parameters in the equivariant linear layer is reduced to $\hat{C} \times N\hat{C}/4$, namely, 25\% of that in the original VMamba. Consequently, the overall network achieves a reduction of more than 50\% in total parameter count, as reported in Table~\ref{tab1:imagenet}.
	
	\subsection{Implementation of Rotation Equivariant Visual Mamba}
	Leveraging the proposed equivariant modules, we transform two representative Mamba-based architectures, VMamba for high-level and mid-level vision tasks and MambaIR for low-level vision tasks, into their equivariant counterparts: EQ-VMamba and EQ-MambaIR.
	Beyond these instantiations, the proposed formulation can be extended to render other Mamba-based visual frameworks end-to-end rotation equivariant.
	
	\textbf{EQ-VMamba for Classification and Segmentation.}
	For high-level and mid-level vision tasks, we first construct an EQ-VMamba backbone for feature extraction by following the basic architecture of VMamba~\cite{2024Vmamba}.
	As illustrated in Fig.~\ref{fig:EQ-VMamba_architecture}, the EQ-VMamba backbone begins with an EQ-patch embedding module that tokenizes the input image into a group-structured feature map.
	Subsequently, four network stages are employed to create hierarchical representations.
	Each stage comprises a stack of EQ-VSS blocks, followed by an EQ-downsampling layer (except for the final stage).
	Furthermore, all remaining non-equivariant modules---including depthwise convolution, LayerNorm2d, and  Dropout---are systematically replaced with their rotation equivariant counterparts. 
	
	{For image classification, the classification head consists of a global average pooling layer over the spatial dimensions, an EQ-Linear layer, and a final averaging over the group dimension, which together produce rotation invariant predictions.}
	For semantic segmentation, we re-engineer the classical UPerNet~\cite{xiao2018unified} decoder into a rotation equivariant counterpart (EQ-UPerNet) by systematically substituting its constituent modules with EQ-CNN, EQ-Dropout, and EQ-upsampling counterparts.
	{The EQ-UPerNet decoder averages along the group dimension only at the last step, so that the segmentation prediction remains rotation \emph{equivariant} and pixel-wise aligned with the rotated input.}
	{Finally, we obtain an end-to-end rotation invariant EQ-VMamba for image classification and a rotation equivariant EQ-VMamba for semantic segmentation.}
	
	\textbf{EQ-MambaIR for Image Restoration.}
	For low-level vision tasks, we develop EQ-MambaIR based on the MambaIR~\cite{2024Mambair} architecture.
	The majority of MambaIR's components, such as VSS blocks and convolutional layers, can be directly replaced with their equivariant counterparts.
	For the reconstruction layers, we replace the original non-equivariant PixelShuffle~\cite{shi2016real} with an EQ-PixelShuffle module to achieve equivariant upsampling.
	Additionally, the channel attention mechanism~\cite{hu2018squeeze} employed  in MambaIR is adapted into a rotation equivariant form by replacing its internal linear layers with EQ-Linear layers.
	The resulting EQ-MambaIR framework provides an end-to-end rotation equivariant solution applicable to a broad range  of image restoration tasks.

	\section{Experimental Results}\label{sec:Experiments}
	
	\begin{table*}[!t]
		\small
		\centering
		\setlength{\tabcolsep}{1pt}
		\renewcommand\arraystretch{0.99}
		\caption{Quantitative comparison of image classification performance on the \textbf{ImageNet-100} validation set.}
		\label{tab1:imagenet}
		\vspace{-2mm}
		\resizebox{0.499\linewidth}{!}{
			\setlength{\tabcolsep}{0.1cm}
			
			\begin{tabular}{l|ccccc}
				\Xhline{1.20pt}
				\multirow{2.5}{*}{Model} &  \multirow{2.5}{*}{Arch.} &  \multirow{2.5}{*}{\makecell{\#Param.}} &  \multirow{2.5}{*}{\makecell{FLOPs}} & \multirow{2.5}{*}{\makecell{Top-1\\(\%)}} &  \multirow{2.5}{*}{\makecell{Top-5\\(\%)}} \\
				& &  &  &   \\[3pt]
				\hline
				ConvNeXt-T~\cite{liu2022convnet}              & CNN    & 29M & 4.5G & 87.06 & 96.70 \\
				DeiT-S~\cite{touvron2021training}             & Trans. & 22M & 4.6G & 78.88 & 93.64 \\
				Swin-T~\cite{liu2021swin}                     & Trans. & 29M & 4.5G & 87.42 & 97.20 \\
				XCiT-S24~\cite{ali2021xcit}                   & Trans. & 26M & 9.2G & 87.14 & 96.98 \\
				Vim-T~\cite{zhu2024ViM}                       & SSM    & 7M  & 1.5G & 82.20 & 95.60 \\
				MSVMamba-M~\cite{shi2024multi}                & SSM    & 12M & 1.5G & 87.56 & 97.76 \\
				SpectralVMamba-T~\cite{2025spectralvmamba}    & SSM    & 21M & 3.9G & 87.86 & 97.25 \\
				VMamba-T~\cite{2024Vmamba}                    & SSM    & 30M & 4.9G & 87.80 & 97.70 \\
				\rowcolor{gray!20}EQ-VMamba-T                 & SSM    & \textbf{10M} & 4.9G & \textbf{88.58} & \textbf{98.14}\\
				\Xhline{1.2pt}
			\end{tabular}
		}
		\hspace{-2mm}
		\resizebox{0.499\linewidth}{!}{
			\setlength{\tabcolsep}{0.1cm}
			\begin{tabular}{l|ccccc}
				\Xhline{1.2pt}
				\multirow{2.5}{*}{Model} &  \multirow{2.5}{*}{Arch.} &  \multirow{2.5}{*}{\makecell{\#Param.}} &  \multirow{2.5}{*}{\makecell{FLOPs}} & \multirow{2.5}{*}{\makecell{Top-1\\(\%)}} &  \multirow{2.5}{*}{\makecell{Top-5\\(\%)}} \\
				& &  &  &   \\[3pt]
				\hline
				ConvNeXt-S~\cite{liu2022convnet}              & CNN    & 50M & 8.7G & 87.54 & 96.96 \\
				DeiT-B~\cite{touvron2021training}             & Trans. & 86M & 17.5G& 78.14 & 92.60 \\
				Swin-S~\cite{liu2021swin}                     & Trans. & 50M & 8.7G & 87.26 & 97.46 \\
				XCiT-M24~\cite{ali2021xcit}                   & Trans. & 84M & 16.2G & 87.72 & 97.36 \\
				Vim-S~\cite{zhu2024ViM}                       & SSM    & 26M & 5.1G & 80.24  & 95.00 \\
				MSVMamba-T~\cite{shi2024multi}                & SSM    & 33M & 4.6G & 88.44 & 97.92 \\
				SpectralVMamba-S~\cite{2025spectralvmamba}    & SSM    & 35M & 6.3G & 88.09 & 97.08 \\
				VMamba-S~\cite{2024Vmamba}                    & SSM    & 50M & 8.7G & 88.32 & 97.82\\
				\rowcolor{gray!20}EQ-VMamba-S                 & SSM    & \textbf{17M} & 8.7G & \textbf{88.70} & \textbf{98.22} \\
				\Xhline{1.2pt}
			\end{tabular}
		}
		\vspace{-5mm}
	\end{table*}
	To validate the effectiveness of the proposed framework, we conduct extensive comparative experiments on four representative vision tasks: image classification, semantic segmentation, classical image super-resolution, and lightweight image super-resolution.
	Beyond standard benchmarks, we further evaluate the classification and segmentation performance of VMamba and EQ-VMamba on rotated datasets to assess robustness to input rotations.
	We also perform equivariance verification experiments to empirically quantify the equivariance errors of EQ-VMamba and EQ-MambaIR.
	Additionally, we compare the proposed equivariant Mamba-based architecture with CNN-based and ViT-based equivariant networks on image super-resolution (please refer to Sec.~S3.4 of the supplementary material).
	Finally, systematic ablation studies are conducted to verify the effectiveness of the end-to-end equivariant design and the proposed group Mamba blocks, to disentangle the architectural equivariance from the effect of rotation augmentation, {and to compare EQ-VMamba with the group-averaged VMamba baseline}.
	
	\subsection{Image Classification}
	\textbf{Experimental Settings.}
	We evaluate the classification performance of EQ-VMamba on the ImageNet-100 dataset, a widely used subset of ImageNet-1K~\cite{imagenet}.
	ImageNet-100 consists of 100 categories, with 130K images for training and 5K images for validation.
	Following the training protocol of VMamba~\cite{2024Vmamba}, all models are trained from scratch using the AdamW optimizer~\cite{loshchilov2017decoupled} for 300 epochs with a cosine learning rate decay schedule, a batch size of 1024, an initial learning rate of 0.001, and a weight decay of 0.05.
	To stabilize training, we apply exponential moving average (EMA)~\cite{polyak1992acceleration} to model parameters.
	During training, input images are resized to $224 \times 224$, and standard data augmentations are employed, including color jittering, random rotation ($\pm 30^\circ$), AutoAugment, random erasing, Mixup, and CutMix.
	Moreover, we develop EQ-VMamba at two scales: tiny (EQ-VMamba-T) and small (EQ-VMamba-S).
	For a fair comparison, we reproduce VMamba~\cite{2024Vmamba} under the same experimental settings.
	
	\textbf{Experimental Results.}
	Table~\ref{tab1:imagenet} presents the comparison results on the ImageNet-100 dataset.
	EQ-VMamba outperforms the non-equivariant VMamba baseline on nearly all settings while utilizing approximately $1/3$ of the parameters.
	Specifically, EQ-VMamba-T achieves a top-1 accuracy of \textbf{88.58}\%, exceeding VMamba-T by \textbf{0.78}\% with a parameter reduction from 30M to 10M, {at the cost of a slightly lower inference throughput}\footnote{{See Sec.~S3.7 of the supplementary for detailed efficiency comparisons.}}.
	Similarly, EQ-VMamba-S achieves \textbf{88.70}\% top-1 accuracy, outperforming VMamba-S by \textbf{0.38}\% with a parameter reduction from 50M to 17M.
	Furthermore, our method outperforms the rotation-invariant Spectral VMamba~\cite{2025spectralvmamba} with fewer learnable parameters.
	These results demonstrate that the integration of rotation equivariance serves as a powerful inductive bias, boosting performance while significantly enhancing parameter efficiency through group-wise weight sharing.

	\begin{figure}[t]
		\centering
		\includegraphics[width=0.49\textwidth]{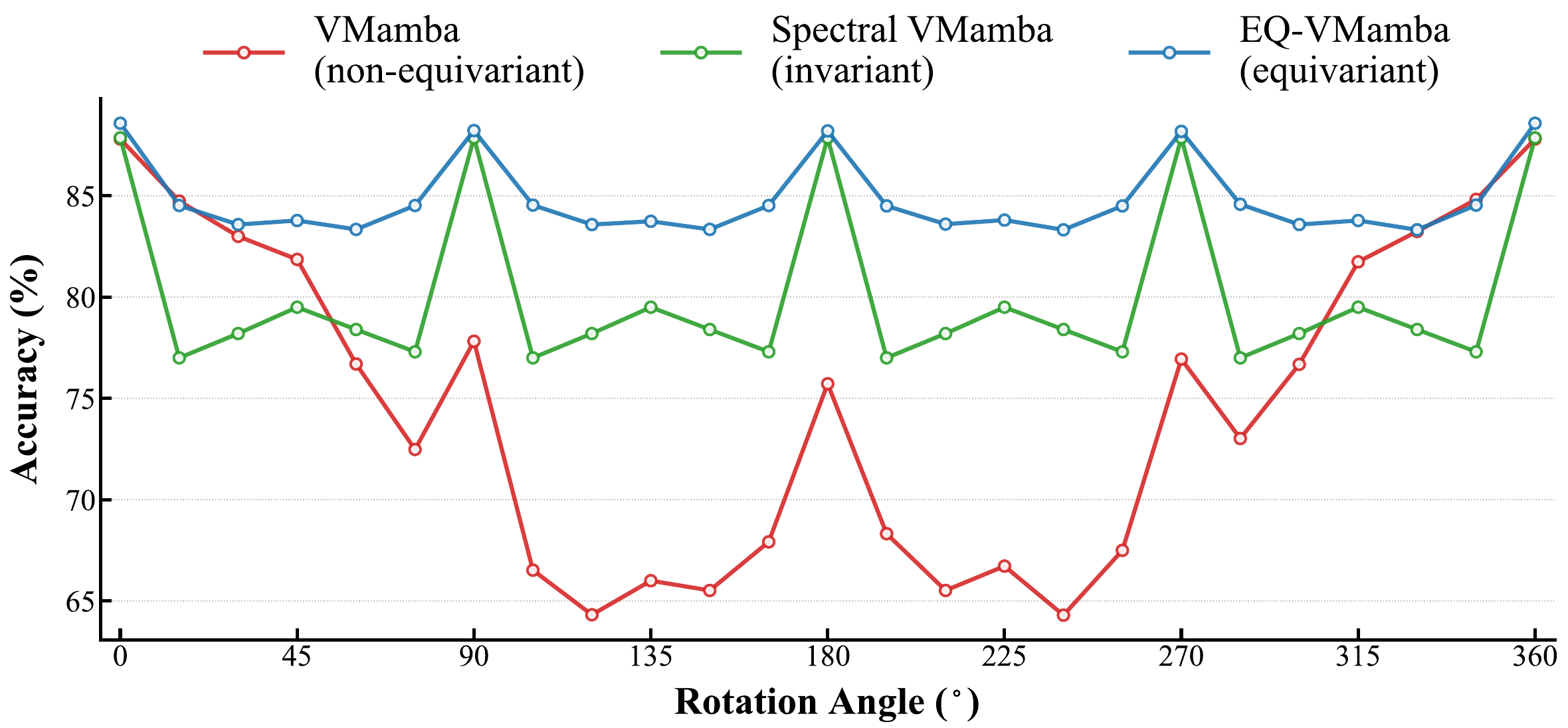}
		\vspace{-7mm}
		\caption{
			Robustness comparison of  VMamba-T, Spectral VMamba-T, and EQ-VMamba-T on the rotated ImageNet-100 dataset.
		}
		\label{fig:visualization_classification}
		\vspace{-2mm}
	\end{figure}

	\textbf{Rotated Classification Experimental Results.}
	Fig.~\ref{fig:visualization_classification} compares the classification robustness of VMamba-T, Spectral VMamba-T, and EQ-VMamba-T under various rotation angles on ImageNet-100. 
	{Test images are rotated from $0^\circ$ to $345^\circ$ with an interval of $15^\circ$ using nearest-neighbor interpolation, and evaluated at their original resolution.}
	The results show that VMamba exhibits pronounced sensitivity to image rotation, suffering from substantial performance degradation.
	While Spectral VMamba maintains stable performance at cardinal rotations (e.g., $0^\circ, 90^\circ, 180^\circ,$ and $270^\circ$), its performance drops sharply at intermediate rotation angles.
	In contrast, EQ-VMamba preserves strict {invariance} under 90-degree rotations and evidently demonstrates superior robustness across the entire rotation spectrum.
	Collectively, these results confirm  that our design successfully alleviates the rotational robustness limitations of VMamba.

	\subsection{Semantic Segmentation} \label{sec:Semantic_Segmentation}
	\textbf{Experimental Settings.}
	We evaluate the semantic segmentation performance of EQ-VMamba on four widely used natural image datasets (ADE20K~\cite{ade20k}, PASCAL VOC 2012~\cite{voc2012}, Cityscapes~\cite{cityscapes}, and COCO-Stuff-164K~\cite{Coco-stuff}), two remote sensing datasets (LoveDA~\cite{LoveDA} and ISPRS Potsdam~\cite{rottensteiner2012isprs}), and {two medical image datasets \footnote{{Additional medical imaging experiments (including pathology classification and CT metal artifact reduction) are reported in Sec. S3.3 of the supplementary material.}} (ISIC 2018~\cite{codella2019skin} and MoNuSAC~\cite{verma2021monusac2020})}.
	All experiments are implemented using the MMSegmentation~\cite{mmsegmentation} framework.
	For Cityscapes and ISIC 2018, input images are cropped to $1024 \times 1024$ with a batch size of 8;
	for all other datasets, input images are cropped to $512 \times 512$ with a batch size of 16.
	For natural images, following prior work~\cite{2025spectralvmamba}, we initialize the EQ-VMamba backbone with ImageNet-100 pretrained weights and adopt EQ-UPerNet as the decoder.
	All models are trained for 160K iterations using the AdamW optimizer~\cite{loshchilov2017decoupled} with an initial learning rate of $6 \times 10^{-5}$.
	For remote sensing and medical image segmentation, following the protocol of Samba~\cite{zhu2024samba}, models are trained from scratch for 15K iterations with an initial learning rate of $6 \times 10^{-4}$.
	To ensure a fair comparison, all comparison methods are reproduced under the same experimental settings.

	\begin{table*}[!t]
		\centering 
		\caption{Quantitative comparison of \textbf{semantic segmentation} performance on natural/remote sensing/medical image datasets (mIoU: \%).}
		\label{tab2:semantic_segmentation_results}
		\vspace{-2mm}
			\setlength{\tabcolsep}{7pt} 
				\begin{tabular}{@{}lccccccc c c@{}}
					\toprule
					\multirow{2.5}{*}{Backbone} & \multirow{2.5}{*}{\makecell{\#Param.}} & \multicolumn{4}{c}{\textbf{Natural Image}} & \multicolumn{2}{c}{\textbf{Remote Sensing Image}}  & 
					\multicolumn{2}{c}{\textbf{Medical Image}}\\
					\cmidrule(lr){3-6} \cmidrule(lr){7-8} \cmidrule(lr){9-10} 
					& & {ADE20K} & {VOC \hspace{-0.5em} 2012} & {Cityscapes} & {COCO-Stuff} & \hspace{0.5em} LoveDA & {Potsdam}  & {ISIC \hspace{-0.5em} 2018} & MoNuSAC  \\
					\midrule
					ConvNeXt-T~\cite{liu2022convnet}            & 48M & 35.59 & 53.99 & 76.33 & 32.81 & \hspace{0.5em} 38.52 & 50.61 & 83.55 & 51.64 \\
					DeiT-S~\cite{touvron2021training}           & 52M & 23.13 & 24.17 & 54.23 & 24.29 & \hspace{0.5em} 38.74 & 48.63 & 83.72 & 39.17 \\
					Swin-T~\cite{liu2021swin}                   & 48M & 27.05 & 35.73 & 70.05 & 25.83 & \hspace{0.5em} 40.78 & 53.78 & 83.73 & 52.02 \\
					XCiT-S24~\cite{ali2021xcit}                 & 76M & 36.21 & 58.08 & 73.02 & 35.11 & \hspace{0.5em} 36.29 & 49.01 & 85.31 & 44.91 \\
					Vim-T~\cite{zhu2024ViM}                     & 13M & 24.31 & 32.20 & 60.04 & 23.64 & \hspace{0.5em} 36.12 & 44.01 & 83.46 & 43.38 \\
					MSVMamba-M~\cite{shi2024multi}              & 42M & 37.69 & 59.78 & 78.40 & 35.94 & \hspace{0.5em} 41.26 & 55.34 & 81.44 & 55.07 \\
					VMamba-T~\cite{2024Vmamba}                  & 62M & \textbf{40.15} & 63.70 & \textbf{78.42} & 38.67 & \hspace{0.5em} 42.62 & 53.24 & 81.75 & 55.93 \\
					\rowcolor{gray!20}   
					EQ-VMamba-T                                 & \textbf{18M} & 39.51 & \textbf{64.46} & 78.01 & \textbf{38.69} & \hspace{0.5em} \textbf{45.69} & \textbf{59.54} & \textbf{85.98} & \textbf{59.28}\\
					
					\midrule
					ConvNeXt-S~\cite{liu2022convnet}            & 70M & 36.89 & 58.83 & 77.12 & 35.03 & \hspace{0.5em} 39.01 & 52.05 & 82.33 & 51.71 \\
					DeiT-B~\cite{touvron2021training}           & 121M & 24.43 & 24.85 & 56.26 & 26.40 & \hspace{0.5em} 39.21 & 50.71 & 81.71 & 39.96 \\
					Swin-S~\cite{liu2021swin}                   & 69M & 29.19 & 38.89 & 73.43 & 28.20 & \hspace{0.5em} 41.23 & 54.69 & 83.34 & 52.29 \\
					XCiT-M24~\cite{ali2021xcit}                 & 112M & 36.39 & 58.63 & 73.84 & 35.75 & \hspace{0.5em} 37.27 & 50.09 & 86.05 & 45.60 \\
					Vim-S~\cite{zhu2024ViM}                     & 46M & 26.62 & 33.16 & 63.48 & 26.91 & \hspace{0.5em} 39.13 & 50.86 & 83.01 & 42.19 \\
					MSVMamba-T~\cite{shi2024multi}              & 65M & 40.70 & 65.98 & 78.35 & 36.63 & \hspace{0.5em} 43.83 & 57.86 & 84.00 & 55.08 \\
					VMamba-S~\cite{2024Vmamba}                  & 82M & \textbf{41.68} & 66.14 & 79.03 &  37.43 & \hspace{0.5em} 40.90 & 56.06 & 83.36 &  56.11 \\
					\rowcolor{gray!20} 
					EQ-VMamba-S                                 & \textbf{25M} & 39.90 & \textbf{66.57} & \textbf{80.36} & \textbf{38.30} & \hspace{0.5em} \textbf{44.04} & \textbf{62.00} & \textbf{86.39} & \textbf{61.05} \\
					\bottomrule
				\end{tabular}
			\vspace{-2mm}
		\end{table*}
		\begin{table*}[!t]
			\centering 
			\caption{Quantitative comparison of \textbf{semantic segmentation} performance on \textbf{rotated} natural/remote sensing/medical datasets (mIoU: \%) (Values inside the parentheses indicate the increase/decrease in performance on rotated inputs compared with unrotated inputs).} 
			\label{tab:rotated_semantic_segmentation_results}
			\vspace{-2mm}
				\setlength{\tabcolsep}{4.0pt} 
				\begin{tabular}{@{}lcccccccc c@{}}
					\toprule
					\multirow{2.5}{*}{Backbone} & \multirow{2.5}{*}{\makecell{\#Param.}} & \multicolumn{4}{c}{\textbf{Natural Image}} & \multicolumn{2}{c}{\textbf{Remote Sensing Image}} & 
					\multicolumn{2}{c}{\textbf{Medical Image}} \\
					\cmidrule(lr){3-6} \cmidrule(lr){7-8} \cmidrule(lr){9-10} 
					& & {ADE20K} & {VOC \hspace{-0.5em} 2012} & {Cityscapes} & {COCO-Stuff} & {LoveDA} & {Potsdam}  & {ISIC \hspace{-0.5em} 2018} & {MoNuSAC}   \\
					\midrule
					VMamba-T~\cite{2024Vmamba} & 62M & 15.98 {\scriptsize\textcolor{deepred}{(24.17$\downarrow$)}}  & 40.42 {\scriptsize\textcolor{deepred}{(23.28$\downarrow$)}} & 32.70 {\scriptsize\textcolor{deepred}{(45.72$\downarrow$)}} & 24.26 {\scriptsize\textcolor{deepred}{(14.41$\downarrow$)}} & 41.68 {\scriptsize\textcolor{deepred}{(0.94$\downarrow$)}} & 53.15 {\scriptsize\textcolor{deepred}{(0.09$\downarrow$)}} & 81.31 {\scriptsize\textcolor{deepred}{(0.44$\downarrow$)}} & 52.74 {\scriptsize\textcolor{deepred}{(3.19$\downarrow$)}}\\
					\rowcolor{gray!20}
					EQ-VMamba-T & \textbf{18M} & \textbf{39.59} {\scriptsize\textcolor{green!50!black}{({0.08}$\uparrow$)}} & \textbf{63.19} {\scriptsize\textcolor{deepred}{({1.27}$\downarrow$)}} & \textbf{78.01} {\scriptsize\textcolor{green!50!black}{({0.00})}} & \textbf{37.80} {\scriptsize\textcolor{deepred}{({0.89}$\downarrow$)}}  & \textbf{45.69} {\scriptsize\textcolor{green!50!black}{({0.00})}} & \textbf{59.54} {\scriptsize\textcolor{green!50!black}{({0.00})}} & \textbf{85.98} {\scriptsize\textcolor{green!50!black}{({0.00})}} & \textbf{60.22} {\scriptsize\textcolor{green!50!black}{(0.94$\uparrow$)}}\\
					\midrule
					VMamba-S~\cite{2024Vmamba} & 82M & 19.60 {\scriptsize\textcolor{deepred}{(22.08$\downarrow$)}}  & 47.54 {\scriptsize\textcolor{deepred}{(18.60$\downarrow$)}} & 39.42 {\scriptsize\textcolor{deepred}{(39.61$\downarrow$)}} & 26.51 {\scriptsize\textcolor{deepred}{(10.92$\downarrow$)}} & 39.19 {\scriptsize\textcolor{deepred}{(1.71$\downarrow$)}} & 53.08 {\scriptsize\textcolor{deepred}{(2.98$\downarrow$)}} & 83.46 {\scriptsize\textcolor{green!50!black}{(0.10$\uparrow$)}} & 56.74 {\scriptsize\textcolor{green!50!black}{(0.63$\uparrow$)}}\\
					\rowcolor{gray!20}
					EQ-VMamba-S & \textbf{25M} & \textbf{39.87} {\scriptsize\textcolor{deepred}{({0.03}$\downarrow$)}} & \textbf{65.64} {\scriptsize\textcolor{deepred}{({0.93}$\downarrow$)}} & \textbf{80.36} {\scriptsize\textcolor{green!50!black}{({0.00})}} & \textbf{37.76} {\scriptsize\textcolor{deepred}{({0.54}$\downarrow$)}}  & \textbf{44.04} {\scriptsize\textcolor{green!50!black}{({0.00})}} & \textbf{62.00} {\scriptsize\textcolor{green!50!black}{({0.00})}} & \textbf{86.39} {\scriptsize\textcolor{green!50!black}{({0.00})}}& \textbf{61.22} {\scriptsize\textcolor{green!50!black}{(0.17$\uparrow$)}}\\
					\bottomrule
				\end{tabular}
			\vspace{-2mm}
		\end{table*}

\begin{figure}[t]
			\centering
			\includegraphics[width=0.5\textwidth]{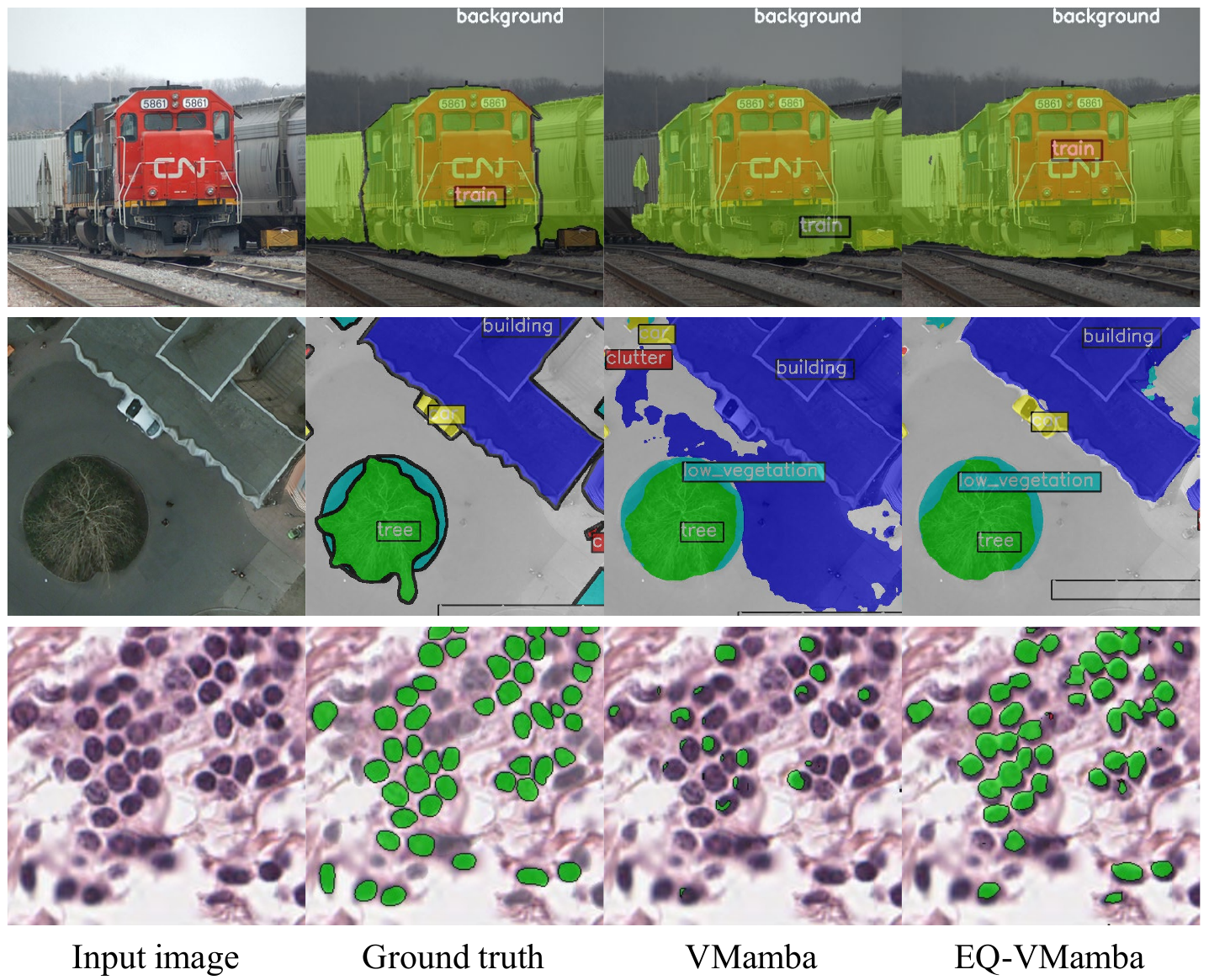}
			\vspace{-7mm}
			\caption{
				Visual comparison of segmentation results between VMamba and EQ-VMamba on PASCAL VOC 2012 (natural image, top), ISPRS Potsdam (remote sensing, middle), and {MoNuSAC (medical image, bottom)}.
			}
			\label{fig:visualization_segmentation}
			\vspace{-4mm}
		\end{figure}

		\textbf{Experimental Results.}
		Table~\ref{tab2:semantic_segmentation_results} summarizes the comparison results across eight semantic segmentation datasets.
		Overall, EQ-VMamba achieves comparable performance to VMamba on natural image datasets while utilizing less than one third of the parameters, and substantially outperforms VMamba on remote sensing datasets.
		Specifically, on natural images, EQ-VMamba-S surpasses VMamba-S on PASCAL VOC 2012, Cityscapes, and COCO-Stuff-164K, and falls behind only on ADE20K.
		The advantages of the proposed equivariant architecture become more pronounced on remote sensing and medical imaging domains.
		On remote sensing data, EQ-VMamba-S exceeds VMamba-S by \textbf{3.14}\% mIoU on LoveDA and \textbf{5.94}\% mIoU on ISPRS Potsdam.
		{On medical images, EQ-VMamba-S continues to improve over VMamba-S by \textbf{3.03}\% mIoU on ISIC 2018 and \textbf{4.94}\% mIoU on MoNuSAC.}
		A similar trend is observed for the tiny variants. 
		Qualitative visualizations in Fig.~\ref{fig:visualization_segmentation} present several examples in which EQ-VMamba produces more accurate segmentation predictions than the original VMamba.

\begin{figure}[t]
			\centering
			\includegraphics[width=0.5\textwidth]{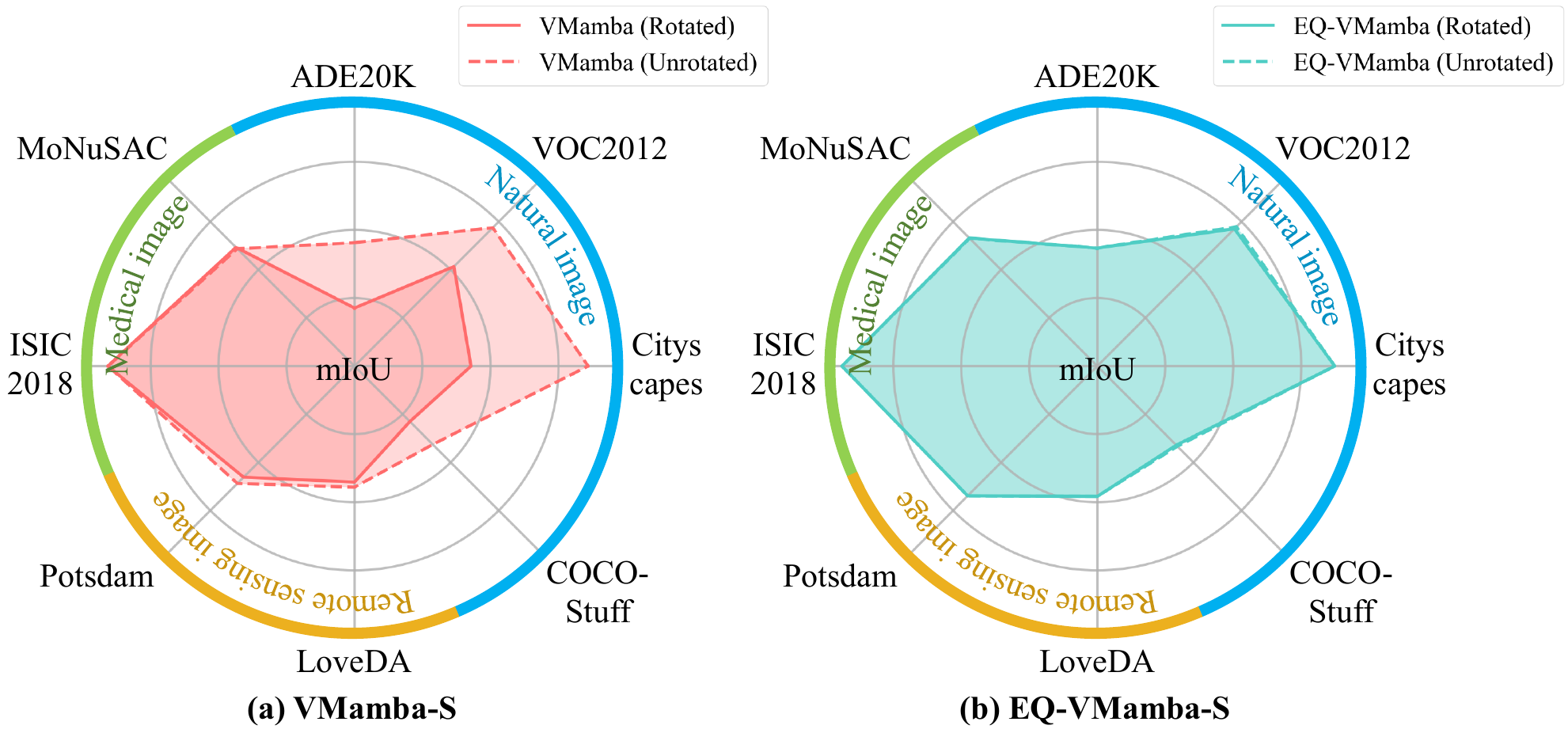}
			\vspace{-6mm}
			\caption{
				Robustness comparison of VMamba and EQ-VMamba on the rotated semantic segmentation datasets.
				Input image rotations lead to a substantial performance degradation in VMamba, while EQ-VMamba preserves performance under 90-degree rotations.
			}
			\label{fig:Rotate_segmentation}
			\vspace{-2mm}
		\end{figure}

		\textbf{Rotated Segmentation Experimental Results.}
		Table~\ref{tab:rotated_semantic_segmentation_results} and Fig.~\ref{fig:Rotate_segmentation} compare the semantic segmentation robustness of VMamba and EQ-VMamba under cardinal rotations ($0^\circ, 90^\circ, 180^\circ, 270^\circ$) across both natural, remote sensing, and medical imaging datasets.
		Experimental results show that the vanilla VMamba suffers substantial performance degradation under rotation on natural images, and marginal degradation on remote sensing images.
		For example, VMamba-T exhibits a drastic segmentation accuracy decline of \textbf{24.17\%} on the rotated ADE20K dataset.
		In contrast, EQ-VMamba maintains nearly constant performance across all datasets, demonstrating markedly superior robustness.
		{All rotated samples are evaluated at their original resolution.
			The small fluctuations observed on certain datasets (e.g., ADE20K) arise because rotating an image whose resolution is not divisible by the overall downsampling factor changes the patch partitioning.}

		\textbf{Analysis of Experimental Results.}
		Experimental results in Tables~\ref{tab2:semantic_segmentation_results} and \ref{tab:rotated_semantic_segmentation_results} reveal a noteworthy phenomenon: EQ-VMamba achieves substantially larger performance gains on remote sensing and medical imaging datasets than natural image datasets.
		Correspondingly, the vanilla VMamba exhibits stronger rotational robustness on remote sensing datasets than natural image datasets.
		{A plausible explanation for both observations lies in the geometric properties of the underlying data distributions.}
		Natural images in these datasets (e.g., ADE20K and Cityscapes) typically adhere to canonical upright orientations and lack global rotational symmetry {(e.g., \textit{sky} typically appears at the top and \textit{road} at the bottom of street scenes)}, which limits the benefits of equivariant architectures.
		In such settings, the parameter reduction induced by group-wise weight sharing may slightly constrain model capacity, leading to marginal performance trade-offs.
		In contrast, remote sensing and medical images---captured from nadir or aerial perspectives---inherently exhibit stronger rotational symmetry.
		This enables EQ-VMamba to fully exploit its equivariant inductive bias, yielding markedly improved performance.
		{This interpretation is consistent with the label-layout asymmetry $\epsilon_Y$ introduced in Sec.~S2.3 of the supplementary material, which is computed from the ground-truth masks: for each class, we estimate the probability that it appears at each position of the image, and $\epsilon_Y$ measures how much these class probability maps change when the images are rotated, i.e., how strongly the spatial layout of the labels depends on the absolute orientation of the image. 
		$\epsilon_Y$ ranges from 0.10 to 0.86 on the four natural image datasets but does not exceed 0.05 on the remote sensing and medical imaging datasets, and the latter are also the datasets on which EQ-VMamba attains larger improvements. We regard $\epsilon_Y$ as an empirical indicator that correlates with the observed gains, rather than as a predictor or a causal explanation (see Sec.~S2.3 of the supplementary material for details).}

		\begin{table*}[!t]
			\centering 
			\caption{Quantitative comparison of different methods on the \textbf{classic image super-resolution} benchmark.}
			\vspace{-2mm}
			\label{tab3:sr_classical_results}
			\setlength{\tabcolsep}{7.6pt} 
			\begin{tabular}{@{}lcccccccccccc@{}}
				\toprule
				\multirow{2.5}{*}{Model} & \multirow{2.5}{*}{Scale} & \multirow{2.5}{*}{\makecell{\#Param.}} & \multicolumn{2}{c}{\textbf{Set5}} & \multicolumn{2}{c}{\textbf{Set14}} & \multicolumn{2}{c}{\textbf{BSD100}} & \multicolumn{2}{c}{\textbf{Urban100}} & \multicolumn{2}{c}{\textbf{Manga109}} \\
				\cmidrule(lr){4-5} \cmidrule(lr){6-7} \cmidrule(lr){8-9} \cmidrule(lr){10-11} \cmidrule(lr){12-13}
				& & & PSNR & SSIM & PSNR & SSIM & PSNR & SSIM & PSNR & SSIM & PSNR & SSIM \\
				\midrule
				EDSR~\cite{lim2017enhanced}      & $\times 2$ & 42.6M & 38.11 & 0.9602 & 33.92 & 0.9195 & 32.32 & 0.9013 & 32.93 & 0.9351 & 39.10 & 0.9773 \\
				RCAN~\cite{zhang2018image}       & $\times 2$ & 15.4M & 38.27 & 0.9614 & 34.12 & 0.9216 & 32.41 & 0.9027 & 33.34 & 0.9384 & 39.44 & 0.9786 \\
				SAN~\cite{dai2019second}         & $\times 2$ & 15.7M & 38.31 & 0.9620 & 34.07 & 0.9213 & 32.42 & 0.9028 & 33.10 & 0.9370 & 39.32 & 0.9792 \\
				HAN~\cite{niu2020single}         & $\times 2$ & 15.9M & 38.27 & 0.9614 & 34.16 & 0.9217 & 32.41 & 0.9027 & 33.35 & 0.9385 & 39.46 & 0.9785 \\
				IPT~\cite{chen2021pre}           & $\times 2$ & 115M & 38.37 & -      & 34.43 & -      & 32.48 & -      & 33.76 & -      & -     & -      \\
				SwinIR~\cite{liang2021swinir}    & $\times 2$ & 11.8M & 38.42 & 0.9623 & 34.46 & 0.9250 & 32.53 & 0.9041 & 33.81 & 0.9427 & 39.92 & 0.9797 \\
				EDT~\cite{li2021efficient}       & $\times 2$ & 11.5M & 38.45 & 0.9624 & 34.57 & 0.9258 & 32.52 & 0.9041 & 33.80 & 0.9425 & 39.93 & 0.9800 \\
				SRFormer~\cite{zhou2023srformer} & $\times 2$ & 10.4M & 38.51 & {0.9627} & 34.44 & 0.9253 & 32.57 & 0.9046 & 34.09 & {0.9449} & 40.07 & 0.9802 \\
				MambaIR~\cite{2024Mambair}       & $\times 2$ & 20.4M & {38.57} & {0.9627} & {34.67} & {0.9261} & {32.58} & {0.9048} & {34.15} & 0.9446 & {40.28} & {0.9806} \\
				\rowcolor{gray!20}
				EQ-MambaIR                       & $\times 2$ & 12.1M & \textbf{38.59} & \textbf{0.9629} & \textbf{34.76} & \textbf{0.9268} & \textbf{32.63} & \textbf{0.9056} & \textbf{34.32} & \textbf{0.9458} & \textbf{40.34} & \textbf{0.9810}	\\
				
				\midrule
				EDSR~\cite{lim2017enhanced}      & $\times 3$ & 42.6M & 34.65 & 0.9280 & 30.52 & 0.8462 & 29.25 & 0.8093 & 28.80 & 0.8653 & 34.17 & 0.9476 \\
				RCAN~\cite{zhang2018image}       & $\times 3$ & 15.4M & 34.74 & 0.9299 & 30.65 & 0.8482 & 29.32 & 0.8111 & 29.09 & 0.8702 & 34.44 & 0.9499 \\
				SAN~\cite{dai2019second}         & $\times 3$ & 15.7M & 34.75 & 0.9300 & 30.59 & 0.8476 & 29.33 & 0.8112 & 28.93 & 0.8671 & 34.30 & 0.9494 \\
				HAN~\cite{niu2020single}         & $\times 3$ & 16.1M & 34.75 & 0.9299 & 30.67 & 0.8483 & 29.32 & 0.8110 & 29.10 & 0.8705 & 34.48 & 0.9500 \\
				IPT~\cite{chen2021pre}           & $\times 3$ & 115M & 34.81 & -      & 30.85 & -      & 29.38 & -      & 29.49 & -      & -     & -      \\
				SwinIR~\cite{liang2021swinir}    & $\times 3$ & 11.8M & 34.97 & 0.9318 & 30.93 & 0.8534 & 29.46 & 0.8145 & 29.75 & 0.8826 & 35.12 & 0.9537 \\
				EDT~\cite{li2021efficient}       & $\times 3$ & 11.5M & 34.97 & 0.9316 & 30.89 & 0.8527 & 29.44 & 0.8142 & 29.72 & 0.8814 & 35.13 & 0.9534 \\
				SRFormer~\cite{zhou2023srformer} & $\times 3$ & 10.6M & 35.02 & {0.9323} & 30.94 & {0.8540} & 29.48 & 0.8156 & {30.04} & {0.8865} & 35.26 & 0.9543 \\
				MambaIR~\cite{2024Mambair}       & $\times 3$ & 20.4M & {35.08} & {0.9323} & {30.99} & 0.8536 & {29.51} & {0.8157} & 29.93 & 0.8841 & {35.43} & {0.9546} \\
				\rowcolor{gray!20}
				EQ-MambaIR                       & $\times 3$ & 12.2M & \textbf{35.13} & \textbf{0.9327} & \textbf{31.08} & \textbf{0.8546} & \textbf{29.53} & \textbf{0.8170} & \textbf{30.06} & \textbf{0.8869} & \textbf{35.48} & \textbf{0.9550} \\
				
				\midrule
				EDSR~\cite{lim2017enhanced}      & $\times 4$ & 43.0M & 32.46 & 0.8968 & 28.80 & 0.7876 & 27.71 & 0.7420 & 26.64 & 0.8033 & 31.02 & 0.9148 \\
				RCAN~\cite{zhang2018image}       & $\times 4$ & 15.6M & 32.63 & 0.9002 & 28.87 & 0.7889 & 27.77 & 0.7436 & 26.82 & 0.8087 & 31.22 & 0.9173 \\
				SAN~\cite{dai2019second}         & $\times 4$ & 15.7M & 32.64 & 0.9003 & 28.92 & 0.7888 & 27.78 & 0.7436 & 26.79 & 0.8068 & 31.18 & 0.9169 \\
				HAN~\cite{niu2020single}         & $\times 4$ & 16.1M & 32.64 & 0.9002 & 28.90 & 0.7890 & 27.80 & 0.7442 & 26.85 & 0.8094 & 31.42 & 0.9177 \\
				IPT~\cite{chen2021pre}          & $\times 4$ & 116M & 32.64 & -      & 29.01 & -      & 27.82 & -      & 27.26 & -      & -      & -     \\
				SwinIR~\cite{liang2021swinir}    & $\times 4$ & 11.9M & 32.92 & 0.9044 & 29.09 & 0.7950 & 27.92 & 0.7489 & 27.45 & 0.8254 & 32.03 & 0.9260 \\
				EDT~\cite{li2021efficient}       & $\times 4$ & 11.6M & 32.82 & 0.9031 & 29.09 & 0.7939 & 27.91 & 0.7483 & 27.46 & 0.8246 & 32.05 & 0.9254 \\
				SRFormer~\cite{zhou2023srformer} & $\times 4$ & 10.5M & 32.93 & 0.9041 & 29.08 & 0.7953 & 27.94 & 0.7502 & {27.68} & \textbf{0.8311} & 32.21 & {0.9271} \\
				MambaIR~\cite{2024Mambair}       & $\times 4$ & 20.4M & {33.03} & {0.9046} & {29.20} & {0.7961} & \textbf{27.98} & {0.7503} & {27.68} & 0.8287 & {32.32} & \textbf{0.9272} \\
				\rowcolor{gray!20}
				EQ-MambaIR                       & $\times 4$ & 12.2M & \textbf{33.04} & \textbf{0.9047} & \textbf{29.21} & \textbf{0.7964} & \textbf{27.98} & \textbf{0.7506} & \textbf{27.70} & {0.8300} & \textbf{32.33} & 0.9270 \\
				\bottomrule
				\vspace{-3mm}
			\end{tabular}
		\end{table*}

		\subsection{Classic Image Super-Resolution}\label{sec:Experiments_classic_SR}
		
		\textbf{Experimental Settings.}
		We evaluate the classic image super-resolution (SR) performance of EQ-MambaIR on standard benchmarks under the bicubic degradation setting.
		Following the training protocol of MambaIR~\cite{2024Mambair}, all models are trained on DIV2K~\cite{timofte2017ntire} and Flickr2K~\cite{lim2017enhanced} datasets, and evaluated on five widely used datasets: Set5~\cite{bevilacqua2012low}, Set14~\cite{zeyde2012single}, BSD100~\cite{martin2001database}, Urban100~\cite{huang2015single}, and Manga109~\cite{matsui2017sketch}.
		During training, we adopt the $L_1$ loss function and optimize the network using AdamW~\cite{loshchilov2017decoupled} with an initial learning rate of  $3 \times 10^{-4}$ and a batch size of 32.
		The input patch size is set to $64 \times 64$.
		For $\times 2$ SR, models are trained from scratch for 500K iterations.
		For $\times 3$ and $\times 4$ SR, models are initialized with the pre-trained $\times 2$ weights and subsequently finetuned for 250K iterations.

		\begin{table*}[!t]
			\centering 
			\caption{Quantitative comparison of different methods on the \textbf{lightweight image super-resolution} benchmark.}
			\vspace{-2mm}
			\label{tab4:sr_lightweight_results}\setlength{\tabcolsep}{7.6pt} 
			\begin{tabular}{@{}lcccccccccccc@{}}
				\toprule
				\multirow{2.5}{*}{Model} & \multirow{2.5}{*}{Scale} & \multirow{2.5}{*}{\makecell{\#Param.}} & \multicolumn{2}{c}{\textbf{Set5}} & \multicolumn{2}{c}{\textbf{Set14}} & \multicolumn{2}{c}{\textbf{BSD100}} & \multicolumn{2}{c}{\textbf{Urban100}} & \multicolumn{2}{c}{\textbf{Manga109}} \\
				\cmidrule(lr){4-5} \cmidrule(lr){6-7} \cmidrule(lr){8-9} \cmidrule(lr){10-11} \cmidrule(lr){12-13}
				& & & PSNR & SSIM & PSNR & SSIM & PSNR & SSIM & PSNR & SSIM & PSNR & SSIM \\
				\midrule
				CARN~\cite{ahn2018fast}                & $\times 2$  & 1,592K & 37.76 & 0.9590 & 33.52 & 0.9166 & 32.09 & 0.8978 & 31.92 & 0.9256 & 38.36 & 0.9765 \\
				IMDN~\cite{hui2019lightweight}         & $\times 2$  & 694K & 38.00 & 0.9605 & 33.63 & 0.9177 & 32.19 & 0.8996 & 32.17 & 0.9283 & 38.88 & 0.9774 \\
				LAPAR-A~\cite{li2020lapar}             & $\times 2$  & 548K & 38.01 & 0.9605 & 33.62 & 0.9183 & 32.19 & 0.8999 & 32.10 & 0.9283 & 38.67 & 0.9772 \\
				LatticeNet~\cite{luo2020latticenet}    & $\times 2$  & 756K & 38.13 & 0.9610 & 33.78 & 0.9193 & 32.25 & 0.9005 & 32.43 & 0.9302 & - & - \\
				SwinIR-light~\cite{liang2021swinir}    & $\times 2$  & 910K & 38.14 & 0.9611 & 33.86 & 0.9206 & 32.31 & 0.9012 & 32.76 & 0.9340 & 39.12 & {0.9783} \\
				ELAN~\cite{zhang2022efficient}         & $\times 2$  & 621K & 38.17 & 0.9611 & 33.94 & 0.9207 & 32.30 & 0.9012 & 32.76 & 0.9340 & 39.11 & 0.9782 \\
				SRFormer-light~\cite{zhou2023srformer} & $\times 2$ & 853K & \textbf{38.23} & \textbf{0.9613} & 33.94 & {0.9209} & {32.36} & {0.9019} & 32.91 & 0.9353 & 39.28 & \textbf{0.9785} \\
				MambaIR-light~\cite{2024Mambair}       & $\times 2$ & 859K & 38.16 & 0.9610 & \textbf{34.00} & \textbf{0.9212} & 32.34 & 0.9017 & {32.92} & {0.9356} & {39.31} & 0.9779 \\
				\rowcolor{gray!20}
				EQ-MambaIR-light                       & $\times 2$ & \textbf{519K} & {38.19} & {0.9612} & {33.99} & 0.9208 & \textbf{32.37} & \textbf{0.9022} & \textbf{33.22} & \textbf{0.9375} & \textbf{39.42} & 0.9782\\
				
				\midrule
				CARN~\cite{ahn2018fast}                & $\times 3$  & 1,592K & 34.29 & 0.9255 & 30.29 & 0.8407 & 29.06 & 0.8034 & 28.06 & 0.8493 & 33.50  & 0.9440 \\ 
				IMDN~\cite{hui2019lightweight}         & $\times 3$  & 703K & 34.36 & 0.9270 & 30.32 & 0.8417 & 29.09 & 0.8046 & 28.17 & 0.8519 & 33.61 & 0.9445 \\ 
				LAPAR-A~\cite{li2020lapar}             & $\times 3$  & 544K & 34.36 & 0.9267 & 30.34 & 0.8421 & 29.11 & 0.8054 & 28.15 & 0.8523 & 33.51 & 0.9441 \\
				LatticeNet~\cite{luo2020latticenet}    & $\times 3$  & 765K & 34.53 & 0.9281 & 30.39 & 0.8424 & 29.15 & 0.8059 & 28.33 & 0.8538 & - & - \\
				SwinIR-light~\cite{liang2021swinir}    & $\times 3$    & 886K & 34.62 & 0.9289 & 30.54 & 0.8463 & 29.20 & 0.8082 & 28.66 & 0.8624 & 33.98 & 0.9478 \\
				ELAN~\cite{zhang2022efficient}         & $\times 3$  & 629K & 34.61 & 0.9288 & 30.55 & 0.8463 & 29.21 & 0.8081 & 28.69 & 0.8624 & 34.00 & 0.9478 \\
				SRFormer-light~\cite{zhou2023srformer} & $\times 3$  & 861K & 34.67 & {0.9296} & 30.57 & 0.8469 & 29.26 & 0.8099 & 28.81 & 0.8655 & 34.19 & 0.9489 \\
				MambaIR-light~\cite{2024Mambair}       & $\times 3$   & 867K & {34.72} & {0.9296} & {30.63} & {0.8475} & {29.29} & {0.8099} & {29.00} & {0.8689} & {34.39} & {0.9495} \\
				\rowcolor{gray!20}
				EQ-MambaIR-light                       & $\times 3$ & \textbf{527K} & \textbf{34.81} & \textbf{0.9302} & \textbf{30.66} & \textbf{0.8483} & \textbf{29.32} & \textbf{0.8105} & \textbf{29.10} & \textbf{0.8708} & \textbf{34.48} & \textbf{0.9499} \\
				
				\midrule
				CARN~\cite{ahn2018fast}                & $\times 4$  & 1,592K & 32.13 & 0.8937 & 28.60 & 0.7806 & 27.58 & 0.7349 & 26.07  & 0.7837 & {30.47} & {0.9084} \\
				IMDN~\cite{hui2019lightweight}         & $\times 4$  & 715K & 32.21 & 0.8948 & 28.58 & 0.7811 & 27.56 & 0.7353 & 26.04 & 0.7838 & 30.45 & 0.9075 \\
				LAPAR-A~\cite{li2020lapar}             & $\times 4$  & 659K & 32.15 & 0.8944 & 28.61 & 0.7818 & 27.61 & 0.7366 & 26.14 & 0.7871 & 30.42 & 0.9074 \\
				LatticeNet~\cite{luo2020latticenet}    & $\times 4$  & 777K & 32.30 & 0.8962 & 28.68 & 0.7830 & 27.62 & 0.7367 & 26.25 & 0.7873 & - & - \\
				SwinIR-light~\cite{liang2021swinir}    & $\times 4$  & 897K & 32.44 & 0.8976 & 28.77 & 0.7858 & 27.69 & 0.7406 & 26.47 & 0.7980 & 30.92 & 0.9151 \\
				ELAN~\cite{zhang2022efficient}         & $\times 4$  & 640K & 32.43 & 0.8975 & 28.78 & 0.7858 & 27.69 & 0.7406 & 26.54 & 0.7982 & 30.92 & 0.9150 \\
				SRFormer-light~\cite{zhou2023srformer} & $\times 4$  & 873K & {32.51} & 0.8988 & 28.82 & 0.7872 & 27.73 & 0.7422 & 26.67 & 0.8032 & 31.17 & 0.9165 \\
				MambaIR-light~\cite{2024Mambair}       & $\times 4$  & 879K & {32.51} & {0.8993} & {28.85} & {0.7876} & {27.75} & {0.7423} & {26.75} & {0.8051} & {31.26} & {0.9175} \\
				\rowcolor{gray!20}
				EQ-MambaIR-light                       & $\times 4$ & \textbf{538K} & \textbf{32.70} & \textbf{0.9006} & \textbf{28.89} & \textbf{0.7884} & \textbf{27.78} & \textbf{0.7438} & \textbf{26.85} & \textbf{0.8082} & \textbf{31.31} & \textbf{0.9184}
				\\
				\bottomrule
				\vspace{-3mm}
			\end{tabular}
		\end{table*}

		\begin{figure*}[t]
			\centering
			\includegraphics[width=0.99\textwidth]{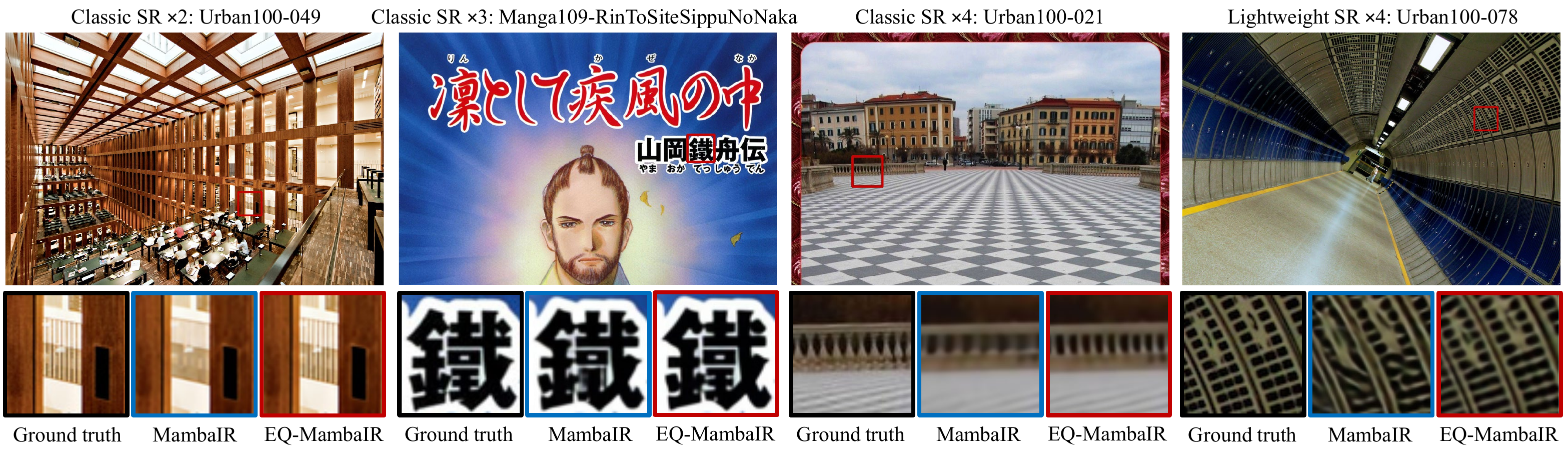}
			\vspace{-3mm}
			\caption{
				Visual comparison of image super-resolution results between MambaIR and EQ-MambaIR on Urban100 and Manga109 datasets.
			}
			\label{fig:visualization_super_resolution}
			\vspace{-2mm}
		\end{figure*}

		\textbf{Experimental Results.}
		Table~\ref{tab3:sr_classical_results} presents the comparison results on standard classic super-resolution benchmarks.
		The proposed EQ-MambaIR outperforms the MambaIR baseline on nearly all test datasets and all upscaling factors ($\times 2$, $\times 3$, and $\times 4$).
		Notably, these improvements are achieved with  approximately half the parameters of the original MambaIR.
		For example, on the challenging Urban100 dataset, EQ-MambaIR (12M parameters) surpasses MambaIR (20M parameters) by \textbf{0.17~dB} and \textbf{0.13~dB} in PSNR at $\times 2$ and $\times 3$ scales, respectively.
		Consistent gains are observed across the remaining test sets, as detailed in the table.
		Additionally, the qualitative comparisons in Fig.~\ref{fig:visualization_super_resolution} further reveal that EQ-MambaIR substantially improves the reconstruction quality of  fine-grained structures and textual elements in degraded images.
		These results collectively underscore the efficacy of embedding rotation symmetry priors into the Mamba framework for low-level vision tasks.

		\subsection{Lightweight Image Super-Resolution}
		
		\textbf{Experimental Settings.}
		We further evaluate EQ-MambaIR under the lightweight SR setting, which imposes more stringent capacity constraints.
		Following the lightweight protocol established in MambaIR~\cite{2024Mambair}, only the DIV2K dataset~\cite{timofte2017ntire} is used for training.
		All other experimental configurations, including optimization strategy and evaluation metrics, remain consistent with the classic SR benchmarks.

		\textbf{Experimental Results.}
		Table~\ref{tab4:sr_lightweight_results} presents the comparison results on standard lightweight SR benchmarks.
		EQ-MambaIR-light outperforms the MambaIR-light baseline on most test datasets and upscaling factors, despite using significantly fewer parameters.
		For example, on the challenging Urban100 dataset, EQ-MambaIR-light (519K parameters) significantly surpasses MambaIR-light (859K parameters) by \textbf{0.30~dB} in PSNR at the $ \times 2$ scale, and by \textbf{0.10~dB} in PSNR at both $ \times 3  $ and $  \times 4$ scales.
		On Manga109, EQ-MambaIR-light further achieves gains of \textbf{0.11~dB} and \textbf{0.09~dB} in PSNR at scales $\times 2$ and $\times 3$, respectively.
		These results further substantiate the benefits of incorporating rotation equivariance into Mamba-based architectures, particularly in resource-constrained scenarios where parameter efficiency is paramount.

		\begin{table*}[!t]
			\centering
			\setlength{\tabcolsep}{1pt}
			\renewcommand\arraystretch{1.0}
			\caption{Average equivariance error of different models under the $\mathrm{p}4$ rotation group (NMSE $\downarrow$).}
			\vspace{-2mm}
			\label{tab:eq_error}
				\setlength{\tabcolsep}{0.25cm}
				\begin{tabular}{lcccc}
					\toprule
					\multirow{2.5}{*}{Model} & \multicolumn{2}{c}{\textbf{Image Classification}} &  \multicolumn{2}{c}{\textbf{Semantic Segmentation}} \\
					\cmidrule(r){2-3}\cmidrule(r){4-5} 
					& Untrained & Trained  & Untrained & Trained   \\
					\midrule  
					VMamba-T~\cite{2024Vmamba} & 0.1721 & 0.4404 & 0.3547 & 0.1958  \\ 
					EQ-VMamba-T & \textbf{0.0003} & \textbf{0.0003} & \textbf{0.0004} & \textbf{0.0002}  \\ 
					\midrule  
					VMamba-S~\cite{2024Vmamba} & 0.2017 & 0.1497 & 0.3617 & 0.1809 \\
					EQ-VMamba-S  & \textbf{0.0003} & \textbf{0.0003} & \textbf{0.0004} & \textbf{0.0002} \\
					\bottomrule  
				\end{tabular}
			\hspace{-0mm}
				\setlength{\tabcolsep}{0.25cm}
				\begin{tabular}{lcccc}
					\toprule
					\multirow{2.5}{*}{Model} & \multicolumn{2}{c}{\textbf{Super-Resolution$\times 2$}} &    \multicolumn{2}{c}{\textbf{Super-Resolution$\times$4}} \\
					\cmidrule(r){2-3} \cmidrule(r){4-5} 
					& Untrained & Trained  & Untrained & Trained   \\
					\midrule  
					MambaIR~\cite{2024Mambair} & 0.1548 & 0.0050 & 0.0912 & 0.0099 \\ 
					EQ-MambaIR & \textbf{8.6e-5} & \textbf{0.0004} & \textbf{4.5e-6} & \textbf{0.0001}  \\ 
					\midrule  
					MambaIR-light~\cite{2024Mambair} & 0.4027 & 0.0061 & 0.5000 & 0.0009 \\
					EQ-MambaIR-light & \textbf{0.0003} & \textbf{0.0004} & \textbf{3.3e-05} & \textbf{6.9e-05} \\
					\bottomrule  
				\end{tabular}
			\vspace{-4mm}
		\end{table*}

		\subsection{Equivariance Verification}
		To empirically verify the theoretical results of equivariance error presented in Sec.~\ref{sec:Method}, we further conduct experiments to quantitatively evaluate the equivariance errors of EQ-VMamba and EQ-MambaIR relative to their non-equivariant counterparts under the $\mathrm{p}4$ rotation group.
		
		\textbf{Experimental Settings.}
		Following the equivariance evaluation protocol established in~\cite{2025Bconv}, we measure equivariance by comparing model predictions on original and transformed inputs.
		Specifically, for an input image ${\bm I}$ and a rotation transformation $\piEQ{R}{G}$ sampled from the $\mathrm{p}4$ group, we feed both ${\bm I}$ and its rotated version $\piEQ{R}{G}({\bm I})$ into the model $\mathcal{F}(\cdot)$ to obtain their respective outputs.
		The Normalized Mean Squared Error (NMSE) between $\mathcal{F}\(\piEQ{R}{G}\({\bm I}\)\)$ and $\piEQ{R}{G}\(\mathcal{F}\({\bm I}\)\)$ serves as the quantitative metric for equivariance error.
		For high-level and mid-level vision tasks, we compare the equivariance errors of VMamba and EQ-VMamba on the ImageNet-100 and ADE20K test datasets.
		For low-level vision tasks, we compare the equivariance errors of MambaIR and EQ-MambaIR on the Urban100 dataset for $\times2$ and $\times4$ image super-resolution.
		Since the equivariance of our architecture is an inherent structural property rather than a learned one, we report errors for both randomly initialized (untrained) and fully converged (trained) models to underscore this training-free characteristic.
		{To ensure that each cardinal rotation acts as an exact permutation of the token grid, inputs are zero-padded to a square resolution divisible by the overall downsampling factor (e.g., $32$ for EQ-VMamba). Full protocol details are provided in Sec.~S3.8 of the supplementary material.}

		\begin{table}[!t]
			\centering
			\setlength{\tabcolsep}{7.5pt}
			\renewcommand\arraystretch{0.99}
			\caption{Ablation study on EQ-cross-scan and EQ-VSS block.}
			\vspace{-2mm}
			\label{tab:ablation_eqcross}
			\begin{tabular}{cccccc}
				\toprule
				\makecell{EQ-cross\\scan} & \makecell{EQ-VSS\\block} &  \makecell{Scale} &    \makecell{\#Param.} & \makecell{Top-1\\(\%)} &  \makecell{Top-5\\(\%)} \\
				\midrule
				\ding{55} & \ding{55}     & Tiny & 30M & 87.80 & 97.70 \\
				\ding{51} & \ding{55}     & Tiny & 27M & 87.88 & 97.54 \\
				\ding{51} & \ding{51}     & Tiny & 10M & \textbf{88.58}  & \textbf{98.14}\\
				\midrule
				\ding{55} & \ding{55}     & Small & 50M & 88.32 & 97.82 \\
				\ding{51} & \ding{55}     & Small & 47M & 87.76 & 97.70 \\
				\ding{51} & \ding{51}     & Small & 17M & \textbf{88.70} & \textbf{98.22} \\
				\Xhline{1.2pt}
			\end{tabular}
			\vspace{-4mm}
		\end{table}

		\textbf{Experimental Results.}
		Table~\ref{tab:eq_error} summarizes the average equivariance error under the $\mathrm{p}4$ rotation group.
		Notably, the equivariance errors of EQ-VMamba and EQ-MambaIR are negligible---effectively approaching zero across all settings---and are several orders of magnitude lower than those of their non-equivariant counterparts.
		These results empirically validate that EQ-VMamba and EQ-MambaIR achieve {strict end-to-end rotation equivariance under the $\mathrm{p}4$ group}.
		Consistent with prior observations in~\cite{2025Bconv}, the measured equivariance error of equivariant networks is not exactly zero in practice.
		This residual error is attributable to floating-point precision limitations in the PyTorch implementation rather than any architectural deficiency.
		Overall, these empirical findings are in strict alignment with our theoretical derivations.

		\subsection{Ablation Study}
		\textbf{Effectiveness of Equivariant Modules. }
		To assess the contribution of the equivariant components, we perform an ablation study on the EQ-cross-scan and EQ-VSS blocks.
		We progressively replace the vanilla cross-scan and VSS blocks in the baseline with their equivariant counterparts and evaluate the impact on ImageNet-100.
		As shown in Table~\ref{tab:ablation_eqcross}, employing EQ-cross-scan in isolation yields no consistent improvements.
		Significant improvements are realized only when all non-equivariant modules are jointly replaced with their equivariant counterparts, demonstrating that partial equivariance is insufficient and holistic architectural equivariance is important.

		\begin{table}[!t]
			\centering
			\setlength{\tabcolsep}{6.5pt}
			\renewcommand\arraystretch{0.99}
			\caption{Ablation study on the group Mamba block.}
			\vspace{-2mm}
			\label{tab:ablation_GroupMambablock}
			\begin{tabular}{ccccc}
				\toprule
				\makecell{Mamba block type} & \makecell{Scale} &    \makecell{\#Param.} & \makecell{Top-1\\(\%)} &  \makecell{Top-5\\(\%)} \\
				\midrule
				Independent Mamba block     & Tiny & 9M & 87.98  & 97.96 \\
				Group Mamba block       & Tiny & 10M & \textbf{88.58}  & \textbf{98.14}\\
				\midrule
				Independent Mamba block     & Small & 15M & 88.04 & 97.98 \\
				Group Mamba block       & Small & 17M & \textbf{88.70} & \textbf{98.22} \\
				\Xhline{1.2pt}
			\end{tabular}
			\vspace{-4mm}
		\end{table}

		\begin{table*}[!t]
			\centering
			\caption{Comparison of EQ-VMamba with the group-averaged VMamba (GA-VMamba) baseline.}
			\vspace{-2mm}
			\label{tab:group_average}
			\renewcommand\arraystretch{0.99}
			\resizebox{0.99\linewidth}{!}{
				\setlength{\tabcolsep}{4pt}
				\begin{tabular}{@{}lccccccccccc@{}}
					\toprule
					\multirow{2.5}{*}{Model} & \multirow{2.5}{*}{Equivariance} & \multicolumn{4}{c}{\textbf{ImageNet-100 Classification}} & \multicolumn{6}{c}{\textbf{Semantic Segmentation (mIoU: \%)}} \\
					\cmidrule(lr){3-6} \cmidrule(lr){7-12}
					& & \#Param. & FLOPs & Top-1 & Top-5 & \#Param. & FLOPs & ADE20K & VOC 2012 & LoveDA & Potsdam \\
					\midrule
					VMamba-T                       & none              & 29.6M & 4.9G  & 87.80 & 97.70 & 61.8M & 236G & \textbf{40.15} & 63.70 & 42.62 & 53.24 \\
					GA-VMamba-T ($C\!=\!48$)           & output averaging  &  7.5M & 5.0G  & 84.50 & 96.18 & 15.5M & 237G & 32.34 & 53.33 & 37.77 & 56.52 \\
					GA-VMamba-T ($C\!=\!56$)           & output averaging  & 10.1M & 6.8G  & 85.46 & 96.60 & 20.9M & 317G &  32.95    &   55.59   & 41.15 & 57.10 \\
					\rowcolor{gray!20}
					EQ-VMamba-T                                       & group equivariance & 10.2M & 4.9G  & \textbf{88.58} & \textbf{98.14} & 18.3M & 246G & 39.51 & \textbf{64.46} & \textbf{45.69} & \textbf{59.54} \\
					\midrule
					VMamba-S                       & none              & 49.5M & 8.7G  & 88.32 & 97.82 & 81.7M & 256G & \textbf{41.68} & 66.14 & 40.90 & 56.06 \\
					GA-VMamba-S ($C\!=\!48$)           & output averaging  & 12.5M & 9.0G  & 86.12 & 96.68 & 20.6M & 258G & 34.51 & 59.34 & 38.72 & 56.77 \\
					GA-VMamba-S ($C\!=\!56$)           & output averaging  & 17.0M & 12.2G & 87.66 & 97.46 & 27.8M & 345G & 36.41 & 60.31 & 41.09 & 57.83 \\
					\rowcolor{gray!20}
					EQ-VMamba-S                    & group equivariance & 17.0M & 8.7G  & \textbf{88.70} & \textbf{98.22} & 25.1M & 275G & 39.90 & \textbf{66.57} & \textbf{44.04} & \textbf{62.00} \\
					\bottomrule
				\end{tabular}
			}
			\vspace{-5mm}
		\end{table*}

		\textbf{Effectiveness of the Group Mamba block. }
		To assess the importance of feature interaction across the group dimension, we conduct an ablation study on ImageNet-100 in which all group Mamba blocks in EQ-VMamba-T and EQ-VMamba-S are replaced with independent Mamba blocks.
		The independent variant employs standard shared linear layers instead of EQ-Linear layers to generate the parameters $\bm{A}$, $\bm{B}$, and $\bm{C}$.
		While still equivariant, this variant lacks inter-group feature interaction.
		As shown in Table~\ref{tab:ablation_GroupMambablock}, the group Mamba block significantly outperforms the independent variant, improving Top-1 accuracy by \textbf{0.60}\% (tiny) and \textbf{0.66}\% (small), respectively.
		These results demonstrate that integrating features across the group dimension provides a more effective inductive bias than independent processing of group components.

		{
			\textbf{Comparison with the group-averaged VMamba baseline.}
			As discussed in Sec.~II-B, exact $\mathrm{p}4$ equivariance can also be imposed on a non-equivariant backbone by group lifting and averaging. To verify that the Mamba-specific design of EQ-VMamba offers a nontrivial advantage over this generic construction, we build a group-averaged VMamba (GA-VMamba) baseline, which applies a shared-weight VMamba to the four cardinal rotations of the input and averages the inversely rotated outputs, with the loss applied to the averaged output. Since GA-VMamba performs four forward passes per input, the channel width $C$ of its backbone is reduced from 96 to 48 to match the FLOPs of EQ-VMamba, or to 56 to match its parameter count. All models are trained with the same protocol.
			As shown in Table~\ref{tab:group_average}, EQ-VMamba consistently outperforms GA-VMamba at both scales. At matched FLOPs, EQ-VMamba surpasses GA-VMamba by 4.08\%/2.58\% Top-1 (tiny/small) on ImageNet-100 and by 3.02--11.13\% mIoU on the four segmentation datasets; at matched parameters, GA-VMamba remains inferior. 
			The gap arises because output averaging leaves the backbone itself non-equivariant and processes the four orientations independently, whereas EQ-VMamba enforces equivariance within every layer and couples the four orientation components in every block.
			These results indicate that the Mamba-specific components of EQ-VMamba provide a substantial advantage over generic group symmetrization at comparable computational budgets.
		}

		{
		\textbf{Effect of Rotation Augmentation.}
		To verify that the performance improvement stems from the architectural equivariance rather than from rotation augmentation, we conduct an ablation study on the rotation augmentation.
		Specifically, we train VMamba and EQ-VMamba on ImageNet-100 under four rotation augmentation regimes: None ($0^\circ$), Weak ($\pm 30^\circ$), Moderate ($\pm 120^\circ$), and Strong ($\pm 180^\circ$).
		As shown in Table~\ref{tab:classification_results_rotation}, even under the strongest augmentation, the equivariance error of VMamba-T remains $0.1791$, and its accuracy still fluctuates across the four cardinal rotations ($85.00\%$--$86.50\%$).
		Moreover, the robustness of VMamba-T gained through augmentation is paid for on unrotated data, with accuracy decreasing from $87.96\%$ to $86.50\%$.
		In contrast, EQ-VMamba-T consistently achieves superior accuracy and rotation robustness ($88.44\%$--$88.58\%$) across all augmentation regimes.
		The complete experimental settings and results are reported in Sec.~S3.5 of the supplementary material.}

		\begin{table}[!t]
			\centering 
			\caption{
				Comparison of VMamba and EQ-VMamba (tiny variants) under different \textbf{rotation augmentation} settings on ImageNet-100.
			}
			\label{tab:classification_results_rotation}
			\vspace{-2mm}
			\renewcommand\arraystretch{0.99}
				\setlength{\tabcolsep}{4pt} 
					\begin{tabular}{@{}lcccccc@{}}
						\toprule
						\multirow{2}{*}{Backbone} 
						& \multirow{2}{*}{\makecell{Rotation\\Aug.}} 
						& \multirow{2}{*}{\makecell{EQ Error\\(NMSE $\downarrow$)}} 
						& \cellcolor{white} Top-1  
						& \multicolumn{3}{c}{Rotated Test }   
						 \\
						\cmidrule(lr){5-7} 
						& &  & \cellcolor{white}(\%) 
						& {${90^{\circ}}$} 
						& {${180^{\circ}}$} 
						& {${270^{\circ}}$} 
						   \\
						\midrule
						VMamba-T~\cite{2024Vmamba} & $0^\circ$          & 0.4501 & 87.96	& 76.44	& 74.86	& 76.84 \\
						EQ-VMamba-T                & $0^\circ$     & \textbf{0.0003} & \textbf{88.54} & \textbf{88.54} & \textbf{88.54} & \textbf{88.54}\\
						VMamba-T~\cite{2024Vmamba} & $\pm 30^\circ$        & 0.4404 & 87.80 & 77.82 & 75.72 & 76.96 \\
						EQ-VMamba-T                & $\pm 30^\circ$      & \textbf{0.0003} & \textbf{88.58} & \textbf{88.58} & \textbf{88.58} & \textbf{88.58}  \\
						VMamba-T~\cite{2024Vmamba} & $\pm 120^\circ$    & 0.1575 & 86.52 & 84.20 & 79.12 & 83.28  \\
						EQ-VMamba-T                & $\pm 120^\circ$  & \textbf{0.0003} & \textbf{88.44} & \textbf{88.44} & \textbf{88.44} & \textbf{88.44} \\
						VMamba-T~\cite{2024Vmamba} & $\pm 180^\circ$        & 0.1791 & 86.50 & 85.34 & 85.18 & 85.00 \\
						EQ-VMamba-T                & $\pm 180^\circ$ & \textbf{0.0003} & \textbf{88.48} & \textbf{88.48} & \textbf{88.48} & \textbf{88.48}  \\
						\bottomrule
					\end{tabular}
				\vspace{-4mm}
			\end{table}
			
			\section{Conclusion}\label{sec:Conclusion}
			In this work, we have introduced EQ-VMamba, the first rotation equivariant Mamba-based architecture designed for vision tasks.
			Motivated by the observed sensitivity of the vanilla VMamba to image rotations, we incorporate rotation equivariance into the Mamba framework through two core innovations: a rotation equivariant cross-scan strategy and the group Mamba blocks.
			We have provided rigorous theoretical analysis demonstrating that the proposed framework guarantees strict end-to-end 90-degree rotation equivariance.
			Extensive experiments across multiple benchmarks---including high-level image classification, mid-level semantic segmentation, and low-level image super-resolution---demonstrate that EQ-VMamba surpasses or matches its non-equivariant counterpart on most evaluated benchmarks, with particularly pronounced gains on datasets without a canonical orientation, such as remote-sensing and medical images.
			
			The exact guarantee established in this work is mainly stated under the $\mathrm{p}4$ rotation group.
			A natural direction for future work is to explore finer-grained rotation groups, such as p8, and investigate how the trade-off between enhanced equivariance and reduced expressiveness affects empirical performance.
			{Another promising direction involves refining the dataset-level symmetry metric.}
			{Our observations in Sec.~\ref{sec:Semantic_Segmentation} suggest that the efficacy of equivariant networks is associated with the symmetry properties of the underlying data distribution. The label-based indicator used in this work is descriptive; a more formal measure that accounts for the joint distribution of images and labels could further clarify when equivariant designs are beneficial.}
			{Moreover, while substantially reducing the parameter count, EQ-VMamba incurs a modest 1.2$\times$ inference slowdown due to the computational overhead of equivariant operations.}
			Given that EQ-VMamba employs cyclic parameter sharing within its EQ-Linear layers, exploring hardware-aware acceleration strategies for EQ-Linear operations represents an important direction to fully leverage the linear-complexity efficiency of the Mamba framework.

			\bibliographystyle{unsrt}
			\bibliography{egbib}
			
		\end{document}